\documentclass[acmsmall]{acmart}

\usepackage{caption}
\usepackage{subcaption} 
\usepackage{amsmath,amsfonts}
\usepackage{algorithmic}
\usepackage{graphicx}
\usepackage{algorithm}
\usepackage{float}
\usepackage{multirow}
\usepackage{latexsym} 
\usepackage{makecell}
\usepackage{hyperref}

\AtBeginDocument{%
  \providecommand\BibTeX{{%
    \normalfont B\kern-0.5em{\scshape i\kern-0.25em b}\kern-0.8em\TeX}}}

\setcopyright{acmcopyright}
\acmJournal{TIST}
\acmYear{2021} \acmVolume{1} \acmNumber{1} \acmArticle{1} \acmMonth{1} \acmPrice{15.00}\acmDOI{10.1145/3465055}

\begin{document}

\title{An Attentive Survey of Attention Models}

\author{Sneha Chaudhari}
\email{snchaudhari@linkedin.com, chaudharissneha@gmail.com}
\orcid{0000-0001-9704-4242}
\affiliation{%
  \institution{LinkedIn Corporation}
  \streetaddress{700 E Middlefield Rd}
  \city{Mountain View}
  \state{California}
  \country{USA}
  \postcode{94043}
}

\author{Varun Mithal}
\affiliation{%
  \institution{LinkedIn Corporation}
  \streetaddress{700 E Middlefield Rd}
  \city{Mountain View}
  \state{California}
  \country{USA}
  \postcode{94043}
}
\email{vamithal@linkedin.com}

\author{Gungor Polatkan}
\affiliation{%
  \institution{LinkedIn Corporation}
  \streetaddress{700 E Middlefield Rd}
  \city{Mountain View}
  \state{California}
  \country{USA}
  \postcode{94043}
}

\author{Rohan Ramanath}
\affiliation{%
  \institution{LinkedIn Corporation}
  \streetaddress{700 E Middlefield Rd}
  \city{Mountain View}
  \state{California}
  \country{USA}
  \postcode{94043}
}

\renewcommand{\shortauthors}{Chaudhari and Mithal, et al.}

\begin{abstract}
  Attention Model has now become an important concept in neural networks that has been researched within diverse application domains. This survey provides a structured and comprehensive overview of the developments in modeling attention. In particular, we propose a taxonomy which groups existing techniques into coherent categories. We review salient neural architectures in which attention has been incorporated, and discuss applications in which modeling attention has shown a significant impact. We also describe how attention has been used to improve the interpretability of neural networks. Finally, we discuss some future research directions in attention. We hope this survey will provide a succinct introduction to attention models and guide practitioners while developing approaches for their applications.
\end{abstract}




\maketitle

\section{Introduction}

Attention Model(AM), first introduced for Machine Translation \cite{DBLP:journals/corr/BahdanauCB14} has now become a predominant concept in neural network literature. Attention has become enormously popular within the Artificial Intelligence(AI) community as an essential component of 
neural architectures for a remarkably large number of applications in Natural Language Processing (NLP) \cite{DBLP:journals/corr/abs-1902-02181}, Speech \cite{DBLP:journals/corr/ChoCB15} and Computer Vision (CV) \cite{wang2016survey}.

\begin{figure}[!ht]
\centering
\includegraphics[width=7cm, height=7cm, keepaspectratio]{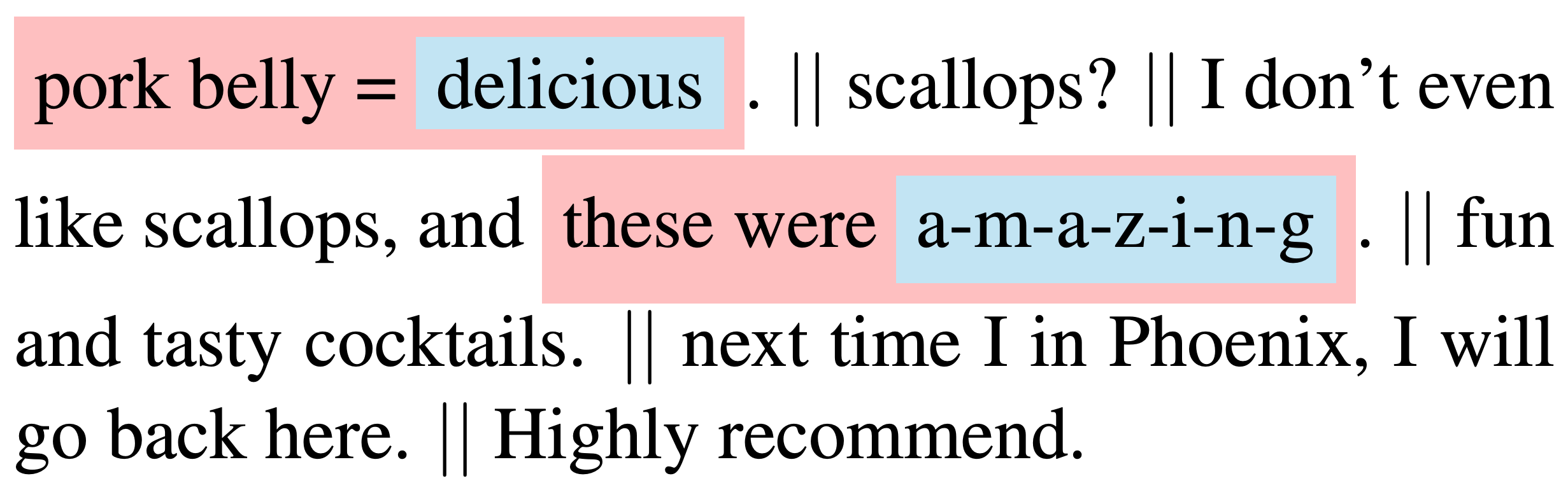} 
\caption{Example of attention modeling in sentiment classification of Yelp reviews. Figure from \cite{Yang2016HierarchicalAN}.}
\label{fig:image0}
\end{figure}

The intuition behind attention can be best explained using human biological systems. For example, our visual processing system tends to focus selectively on some parts of the image, while ignoring other irrelevant information in a manner that can assist in perception \cite{conf/icml/XuBKCCSZB15}. Similarly, in several problems involving language, speech or vision, some parts of the input are more important than others. For instance, in machine translation and summarization tasks, only certain words in the input sequence may be relevant for predicting the next word. Likewise, in an image captioning problem, some regions of the input image may be more relevant for generating the next word in the caption. AM incorporates this notion of relevance by allowing the model to dynamically \textit{pay attention to} only certain parts of the input that help in performing the task at hand effectively. An example of sentiment classification of Yelp reviews \cite{Yang2016HierarchicalAN} using AM is shown in Figure \ref{fig:image0}. In this example, the AM learns that out of five sentences, the first and third sentences are more relevant. Furthermore, the words \textit{delicious} and \textit{amazing} within those sentences are more meaningful to determine the sentiment of the review. 

The rapid advancement in modeling attention in neural networks is primarily due to three reasons. First, these models are now the state-of-the-art for multiple tasks in NLP (Machine Translation, Summarization, Sentiment Analysis, and Part-of-Speech tagging) \cite{DBLP:journals/corr/abs-1902-02181}, Computer Vision (Image Classification, Object Detection, Image Generation) \cite{khan2021transformers}, cross-modal tasks (Multimedia Description, Visual Question Answering) \cite{DBLP:journals/corr/abs-1708-02709} and Recommender Systems \cite{10.1145/3285029}. Second, they offer several other advantages beyond improving performance on the main task. They have been extensively used for improving interpretability of neural networks, which are otherwise considered as black-box models. This is a notable benefit mainly because of growing interest in the fairness, accountability, and transparency of Machine Learning models in applications that influence human lives. Third, they help overcome some challenges with Recurrent Neural Networks(RNNs) such as performance degradation with increase in length of the input and the computational inefficiencies resulting from sequential processing of input (Section \ref{label:am}).

\textit{Organization}: In this work we aim to provide a brief, yet comprehensive survey on attention modeling. In Section \ref{sec:history} we build the intuition for the concept of attention using a simple regression model. We briefly explain the AM proposed by \cite{DBLP:journals/corr/BahdanauCB14} and other attention functions in Section \ref{label:am} and describe our taxonomy in Section \ref{label:taxonomy}. We then discuss key neural architectures using AM and present applications where attention has been widely applied in Section \ref{label:architecture} and \ref{label:applications} respectively. Finally, we describe how attention is facilitating the interpretability of neural networks in Section \ref{label:interpretability} and conclude the paper with future research directions in Section \ref{label:conclusion}.

\textit{Related surveys}: There have been a few domain-specific surveys on attention focusing on Computer Vision \cite{wang2016survey}, and graphs \cite{lee2018attention} and Natural Language Processing \cite{DBLP:journals/corr/abs-1902-02181}. However, we further incorporate an accessible taxonomy, key architectures and applications, and interpretability aspect of AM. We hope that our contributions will not only foster broader understanding of AM but also help AI developers \& engineers to determine the right approach for their application domain.

%which are often not widely available (e.g., book chapters) and targeted at a specialized audience. We give a general survey of SSGs for an AI audience that focuses on recent developments, distills the underlying concepts across the varied applications of SSGs and provides a centralized presentation of open problems in this area. We hope that this introduction to the state-of-the-art will aid new researchers in this area.

\section {Attention Basics} \label{sec:history}
The idea of attention can be understood using a regression model proposed by Naradaya-Watson in 1964 \cite{nadaraya1964estimating, watson1964smooth}. We are given a training data of \emph{n} instances comprising features and their corresponding target values \(\displaystyle \{(x_1, y_1), (x_2, y_2), ..., (x_n, y_n) \} \). We want to predict the target value \(\displaystyle \hat{y} \) for a new query instance \(\displaystyle x \). A naive estimator will predict the simple average of target values of all training instances:  \(\displaystyle \hat{y} = \frac{1}{n} \sum_{i=1}^{n} y_i \). Naradaya-Watson proposed a better approach in which the estimator uses a weighted average where weights correspond to relevance of the training instance to the query: \(\displaystyle \hat{y} = \sum_{i=1}^{n} \alpha(x, x_i) y_i \). Here weighting function \(\displaystyle \alpha(x, x_i) \) \emph{encodes the relevance of instance \(\displaystyle  x_i \) to predict for \(\displaystyle x \)}. A common choice for the weighting function is a normalized Gaussian kernel, though other similarity measures can also be used with normalization. The authors showed that the estimator has (i) consistency: given enough training data it converges to optimal results, and (ii) simplicity: no free parameters, the information is in the data and not in the weights. Fast forward 50 years, attention mechanism in deep models can be viewed as a generalization that also allows learning the weighting function.

\section{Attention Model}
\label{label:am}
\begin{figure}[!h]
\centering
\includegraphics[width=12cm, height=10cm]{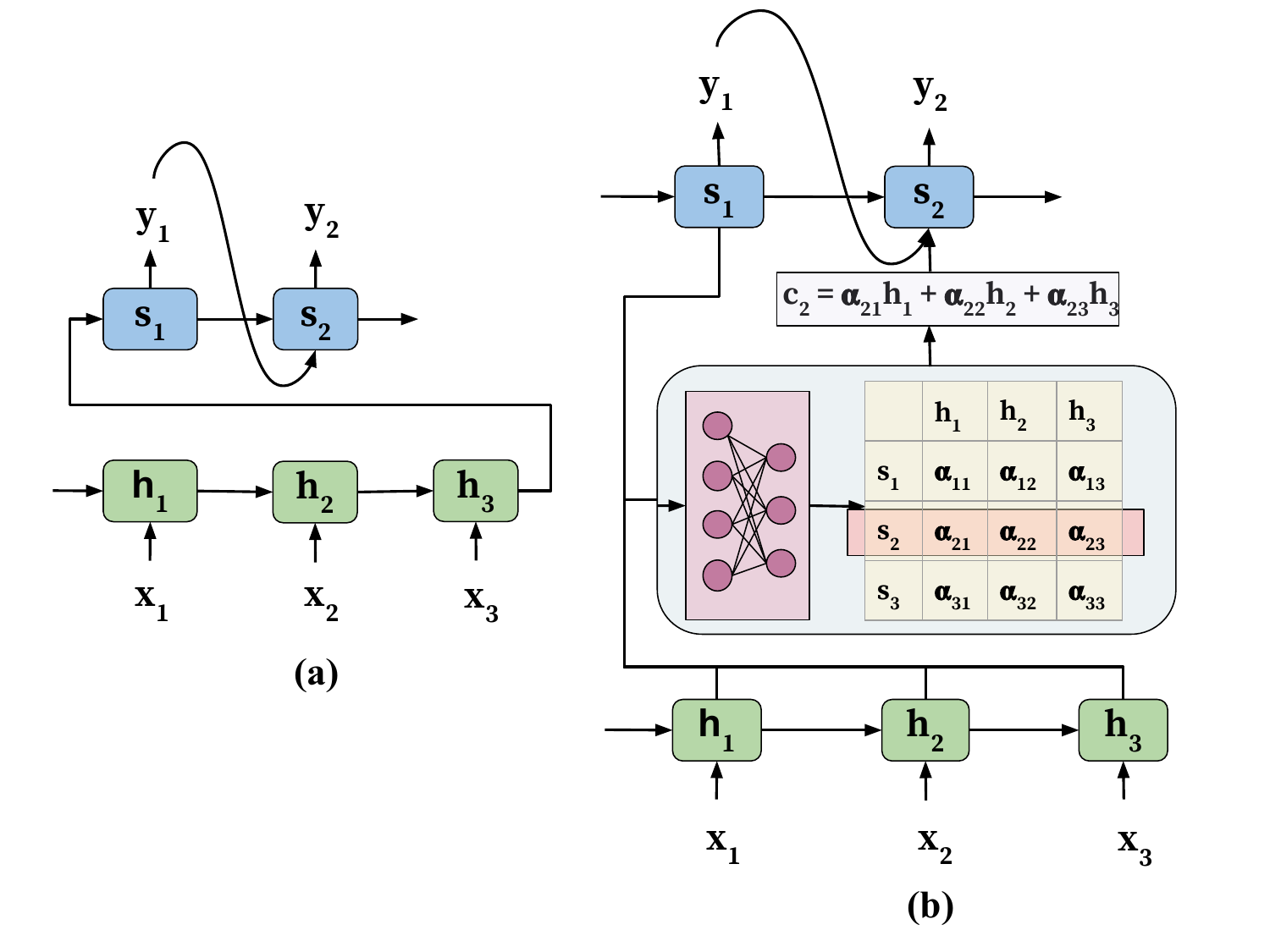} 
\caption{Encoder-decoder architecture: (a) traditional (b) with attention model}
\label{fig:image1}
\vspace*{0.5 cm}
\end{figure}

The first use of AM was proposed by \cite{DBLP:journals/corr/BahdanauCB14} for a sequence-to-sequence modeling task.
A sequence-to-sequence model consists of an encoder-decoder architecture \cite{cho2014learning} as shown in Figure \ref{fig:image1}(a). The encoder is an RNN that takes an input sequence of tokens \(\displaystyle \{x_1, x_2, ..., x_T \} \), where \(\displaystyle T \) is the length of input sequence, and encodes it into fixed length vectors \(\displaystyle \{h_1, h_2, ..., h_T \} \). The decoder is also an RNN which then takes a single fixed length vector \(\displaystyle h_T \) as its input and generates an output sequence \(\displaystyle \{y_1, y_2, ..., y_{T'} \} \) token by token, where \(\displaystyle T' \) is the length of output sequence. At each position \(\displaystyle t \), \(\displaystyle h_t \) and \(\displaystyle s_t \) denote the hidden states of the encoder and decoder respectively.
%The traditional encoder decoder architecture and the one with attention model is shown in  and Figure \ref{fig:image1}(b) respectively. As shown in Figure \ref{fig:image1}(a), encoder receives an input sequence \textit{das grüne haus} and outputs fixed length vector \(\displaystyle h_3 \). The decoder receives \(\displaystyle h_3 \) as input and generates \textit{the green house} sequence as output. Intuitively, at each time step \(\displaystyle t \), \(\displaystyle h_t \) and \(\displaystyle s_t \) denote the hidden states of the encoder and decoder respectively.

\textbf{Challenges of traditional encoder-decoder}: There are two well known challenges with this traditional encoder-decoder framework. First, the encoder has to compress all the input information into a single fixed length vector \(\displaystyle h_T \) that is passed to the decoder. Using a single fixed length vector to compress long and detailed input sequences may lead to loss of information \cite{cho2014properties}. Second, it is unable to model alignment between input and output sequences, which is an essential aspect of structured output tasks such as translation or summarization \cite{DBLP:journals/corr/abs-1708-02709}. Intuitively, in sequence-to-sequence tasks, each output token is expected to be more influenced by some specific parts of the input sequence. However, decoder lacks any mechanism to selectively focus on relevant input tokens while generating each output token. 

\textbf{Key idea}: AM aims at mitigating these challenges by allowing the decoder to access the entire encoded input sequence \(\displaystyle \{h_1, h_2, ..., h_T \} \). The central idea is to induce attention weights \(\displaystyle \alpha \) over the input sequence to prioritize the set of positions where relevant information is present for generating the next output token.

\begin{table}[!h]
\centering
\begin{tabular}{lll}
\hline
Function  & Traditional Encoder- & Encoder-Decoder \\
& Decoder & with Attention  \\
\hline \hline
Encode & \(\displaystyle h_i = f(x_i, h_{i-1}) \) & \(\displaystyle h_i = f(x_i, h_{i-1}) \) \\
Context & \(\displaystyle c = h_T \) & \(\displaystyle c_j = \sum_{i=1}^{T} \alpha_{ij} h_i \) \\
&  & \(\displaystyle \alpha_{ij} = p(e_{ij}) \) \\
& & \(\displaystyle e_{ij} = a(s_{j-1}, h_i) \)\\
Decode & \(\displaystyle s_j = f(s_{j-1}, y_{j-1}, c) \) & \(\displaystyle s_j = f(s_{j-1}, y_{j-1}, c_j) \) \\ 
Generate & \(\displaystyle y_j = g(y_{j-1}, s_j, c) \) & \(\displaystyle y_j = g(y_{j-1}, s_j, c_j) \) \\
\hline \hline
\end{tabular}
\begin{tabular}{p{8 cm}}
\(\displaystyle x = (x_1, ..., x_T) \): input sequence, \(\displaystyle T \): length of input sequence, \(\displaystyle h_{i} \): hidden states of encoder, \(\displaystyle c \): context vector, \(\displaystyle \alpha_{ij} \): attention weights over input, \(\displaystyle s_j \): decoder hidden state, \(\displaystyle y_j \): output token, \(\displaystyle f, g \): non-linear functions, \(\displaystyle a \): alignment function, \(\displaystyle p \): distribution function \\
\hline
\end{tabular}
\vspace*{0.5 cm}
\caption{Encoder-decoder architecture: traditional and with attention model}
\label{tab:math}
\end{table}

\textbf{Usage of attention}: The corresponding encoder-decoder architecture with attention is shown in Figure \ref{fig:image1}(b). The attention block in the architecture is responsible for automatically learning the attention weights \(\displaystyle \alpha_{ij} \), which capture the relevance between \(\displaystyle h_i \) (the encoder hidden state) and \(\displaystyle s_{j-1} \) (the decoder hidden state). Note that the query state \(\displaystyle s_{j-1} \) is hidden state of the decoder just before emitting \(\displaystyle s_{j} \) and \(\displaystyle y_{j} \). These attention weights are then used for building a context vector \(\displaystyle c \), which is passed as an input to the decoder. At each decoding position \(\displaystyle j \), the context vector \(\displaystyle c_j \) is a weighted sum of all hidden states of the encoder and their corresponding attention weights, i.e. \(\displaystyle c_j = \sum_{i=1}^{T} \alpha_{ij}h_i \). This additional context vector is the mechanism by which decoder can access the entire input sequence and also focus on the relevant positions in the input sequence. This not only leads to improvements in performance on the final task but also improves the quality of the output due to better alignment. The same concept is shown mathematically in Table \ref{tab:math}. The only major difference in the encoder-decoder architecture with attention is the composition of context vector \(\displaystyle c \). In the traditional framework, context vector is just the last hidden state of the encoder \(\displaystyle h_T \). In the attention based framework, context at a given decoding step \(\displaystyle j \) is combination of all hidden states of the encoder and their corresponding attention weights; \(\displaystyle c_j = \sum_{i=1}^{T} \alpha_{ij}h_i \).  

% \begin{figure}[!h]
% \centering
% \includegraphics[width=7cm, height=5cm]{math.png} 
% \caption{Encoder-decoder architecture: Vanilla and with attention model (Sneha: change c to $c_j$)}
% \label{fig:imagemath}
% \end{figure}

\textbf{Learning attention weights}: The attention weights are learned by incorporating an additional feed forward neural network within the architecture. This feed forward network learns a particular attention weight \(\displaystyle \alpha_{ij} \) as a function of two states, \(\displaystyle h_i \) (encoder hidden state) and \(\displaystyle s_{j-1} \) (decoder hidden state) which are taken as input by the neural network. This function is called the alignment function (denoted by \emph{a} in Table \ref{tab:math}) as it scores how relevant is the encoder hidden state \(\displaystyle h_i \) for the decoder hidden state \(\displaystyle s_{j-1} \). This alignment function outputs energy scores \(\displaystyle e_{ij} \) which are then fed into the distribution function (denoted by \emph{p} in Table \ref{tab:math}) which converts the energy scores into attention weights. When the functions \(\displaystyle a \) and \(\displaystyle p \) are differentiable, the whole attention based encoder-decoder model becomes one large differentiable function and can be trained jointly with encoder-decoder components of the architecture using simple backpropagation.

\textbf{Generalized Attention Model}: The attention model shown in Figure \ref{fig:image1}(b) can also be seen as a mapping of sequence of keys \(\displaystyle K \) to an attention distribution \(\displaystyle \alpha \) according to query \(\displaystyle q \) where keys are encoder hidden states \(\displaystyle h_i \) and query is the single decoder hidden state \(\displaystyle s_{j-1} \). Here the attention distribution \(\displaystyle \alpha_{ij} \) emphasizes the keys which are relevant for the main task with respect to the query \(\displaystyle q \). Then \(\displaystyle e = a(K, q) \) and \(\displaystyle \alpha = p(e) \). In some cases, there is also additional input of values \(\displaystyle V \) on which the attention distribution is applied. The keys and values generally have one to one mapping and although the core attention model proposed by \cite{DBLP:journals/corr/BahdanauCB14} does not distinguish between keys and values \(\displaystyle (k_i = v_i = h_i) \), some existing literature uses this terminology for different representations of the same input data. Hence a generalized attention model \(\displaystyle A \) works with a set of key-value pairs \(\displaystyle (K, V) \) and query \(\displaystyle q \) such that:

\begin{equation}
A(q,K,V) = \sum_i p(a(k_i, q)) * v_i
\end{equation}

As a concrete example, one can look at the regression task estimator explained in Section \ref{sec:history}. Here the instance \(\displaystyle x \) is the query, the training data points \(\displaystyle x_i \) are keys and their labels \(\displaystyle y_i \) are values.

The alignment function (denoted by \emph{a}) and distribution function (denoted by \emph{p}) determine how keys and query are combined to produce attention weights. We discuss some of the commonly used alignment functions and distribution functions in the literature; we refer the reader to \cite{DBLP:journals/corr/abs-1902-02181} for a more detailed discussion on alignment and distribution functions.

\textbf{Alignment functions} 
The first major category of alignment functions are based on a notion of comparing query representations with key representations. For example, one approach is to compute either the \emph{cosine similarity} or the \emph{dot product} between the key and query representations (see Table \ref{tab:alignmentfn}). To account for varying lengths of representation, \emph{scaled dot product} can be employed that normalizes the dot product by the representation vector length. Note that these functions assume that key and query have the same representation vector space. \emph{General alignment} extends dot product to keys and queries with different representations by introducing a learnable transformation matrix \textbf{W} that maps queries to the vector space of keys. \emph{Biased general alignment} allows to learn the global importance of some keys irrespective of the query by introducing a bias term. \emph{Activated general alignment} adds a nonlinear activation layer such as hyperbolic tangent, rectifier linear unit, or scaled exponential linear unit. More recently, \cite{choromanski2020rethinking} show that key and query can be matched using a generalized kernel function instead of the more commonly used \emph{dot product}. The formulations of these alignment functions are presented in the Table \ref{tab:alignmentfn}.

The second major category of alignment functions combine keys and query to form a joint representation. One of the simplest models that follow this approach is the \emph{concat alignment} by \cite{luong2015effective}, where a joint representation is given by concatenating keys and queries. \emph{Additive alignment} reduces computational time by decoupling the contributions of the query and the key; this allows precomputing contributions of all keys to avoid re-computation for each query. In contrast to a single neural layer used in additive alignment, \emph{deep alignment} employs multiple neural layers.

There are some alignment functions that are designed for specific use-cases. \emph{Location-based alignment} ignores the keys and only depends on q. The alignment score associated with each key is thus computed as a function of the key’s position, independently of its content. \cite{li2019area} show that when working with group of items such as 2-D patches for images or 1-D temporal sequences, \emph{derived features} (such as mean and standard deviation) from the representations of the individual elements belonging to the group can be used as input to alignment functions (such as additive alignment used in the paper).

\begin{table}[!h]
\centering
\begin{tabular}{lll}
\hline
Function  & Equation & References  \\
\hline \hline
similarity & \(\displaystyle a(k_i, q) = sim(k_i, q) \) & \cite{graves2014neural} \\
dot product & \(\displaystyle a(k_i, q) = q^{T}k_i \) & \cite{luong2015effective} \\
scaled dot product & \(\displaystyle a(k_i, q) = \frac{q^{T}k_i}{\sqrt{d_{k}}} \) & \cite{DBLP:journals/corr/VaswaniSPUJGKP17} \\
general & \(\displaystyle a(k_i, q) = q^{T}Wk_i \) &  \cite{luong2015effective} \\
biased general & \(\displaystyle a(k_i, q) = k_i(Wq + b) \) & \cite{sordoni2016iterative} \\
activated general & \(\displaystyle a(k_i, q) = act(q^{T}Wk_i + b) \) & \cite{ma2017interactive} \\
generalized kernel & \(\displaystyle a(k_i, q) = \phi(q)^{T}\phi(k_i) \) & \cite{choromanski2020rethinking} \\
concat & \(\displaystyle a(k_i, q) = w_{imp}^{T}act(W[q;k_i] + b) \) & \cite{luong2015effective} \\
additive & \(\displaystyle a(k_i, q) = w_{imp}^{T}act(W_1q + W_2k_i + b) \) & \cite{DBLP:journals/corr/BahdanauCB14} \\
deep & \(\displaystyle a(k_i, q) = w_{imp}^{T}E^{(L-1)} + b^{L} \) &  \cite{pavlopoulos2017deeper}\\
 & \(\displaystyle  E^{(l)} = act(W_lE^{(l-1)} + b^l  \)   & \\
 & \(\displaystyle E^{(1)} = act(W_1k_i + W_0q) + b^l \) & \\
location-based & \(\displaystyle a(k_i, q) =  a(q) \) &  \cite{luong2015effective} \\
feature-based & \(\displaystyle a(k_i, q) = w_{imp}^{T}act(W_1\phi_1(K) + W_2\phi_2(K) + b) \)  & \cite{li2019area} \\
\hline \hline
\end{tabular}
\begin{tabular}{p{14 cm}}
\(\displaystyle a(k_i,q) \): alignment function for query \emph{q} and key \emph{ \(\displaystyle k_i \)}, sim : similarity functions such as cosine, \(\displaystyle d_k \): length of input, \(\displaystyle (W, w_{imp}, W_0, W_1, W_2) \): trainable parameters, b: trainable bias term, act: activation function. 
\end{tabular}
\vspace*{0.5 cm}
\caption{Summary of Alignment Functions}
\label{tab:alignmentfn}
\end{table}

\textbf{Distribution functions}
Distribution functions map alignment function scores to attention weights. 
The most commonly used distribution functions are \emph{logistic sigmoid} and \emph{softmax}. These functions ensure that attention weights are constrained in [0,1] and sum to 1. Such weights can thus be interpreted as probabilities that an element is relevant. In case of softmax function, attention weights can be interpreted as the probability that the corresponding element is the most relevant. 
Most variants employ a softmax transformation in their attention mechanism, leading to dense alignments. This density is wasteful, making models less interpretable and assigning probability mass to many implausible outputs. Distribution functions such as \emph{sparsemax} \cite{martins2016softmax} and \emph{sparse entmax} \cite{peters2019sparse, martins2020sparse} are able to produce sparse alignments and assign nonzero probability to only a short list of plausible outputs. Sparse distributions could be especially useful in applications such as document summarization or question-answering tasks where a large number of elements are irrelevant.
Finally, compositional de-attention networks \cite{tay2019compositional} introduced a distribution function which forms quasi-attention by using elementwise multiplication of two terms: \(\displaystyle tanh(\frac{qk_i^{T}}{\sqrt{d_k}}) \) and \(\displaystyle sigmoid(\frac{G(qk_i^{T})}{\sqrt{d_k}}) \)(where G(.) is the negation of outer L1 distance between q against all keys). In this case, the first term controls the adding and subtracting of vectors. This is in contrast to traditional attention that only adds (weighted) vectors. The secondary term can be interpreted as a type of gating mechanism that deletes tokens that are irrelevant for the query (by making their contribution zero).

\begin{figure}[!ht]
\centering
\includegraphics[width=11cm, height=9cm, keepaspectratio]{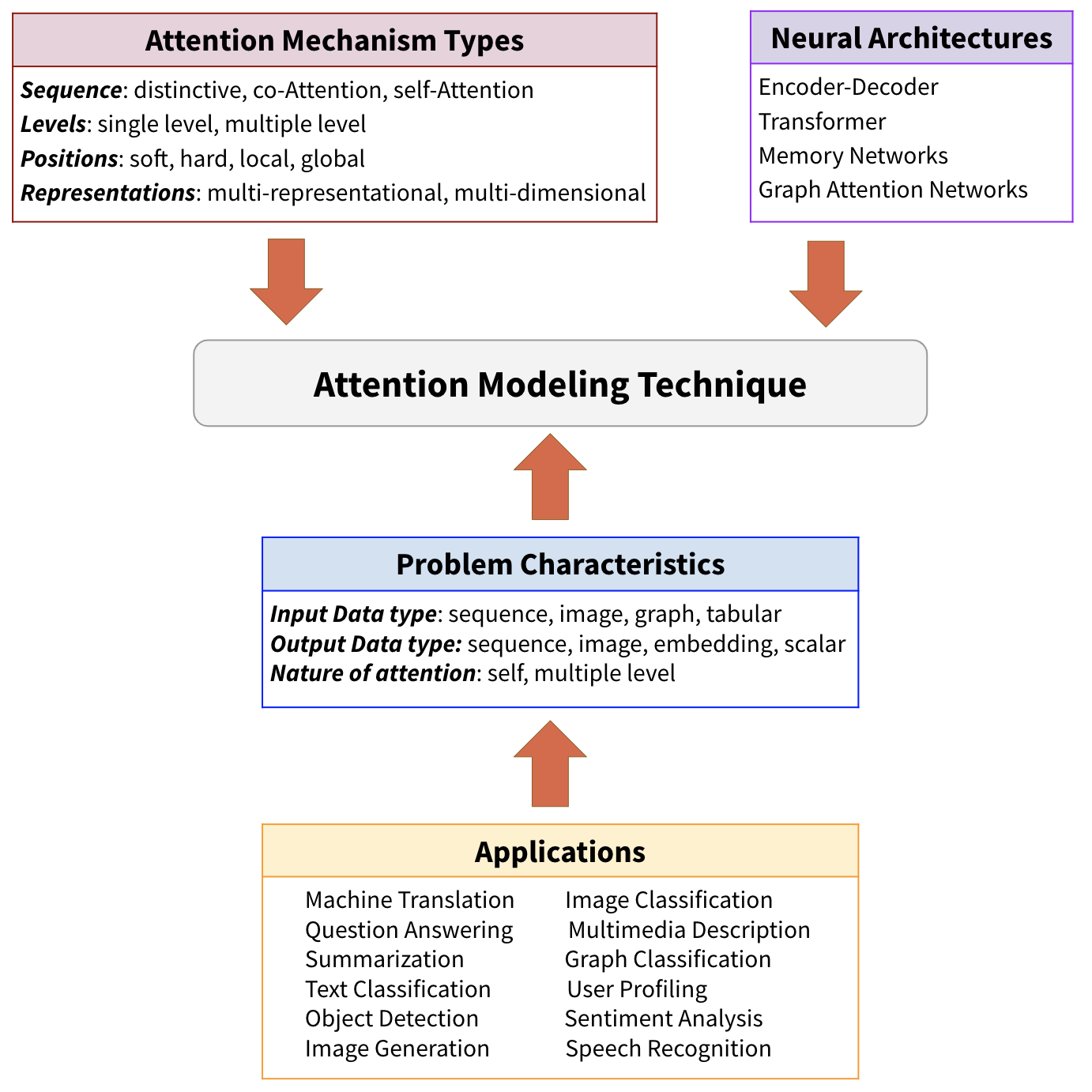}
\caption{Key components of Attention Modeling Techniques}
\label{fig:attmodel}
\end{figure}

In this section we discussed the seminal model that proposed attention mechanism for a sequence-to-sequence task in an encoder-decoder architecture. While the core idea remains the same, several extensions of attention modeling have been proposed in the literature to solve specific problem formulations. We also discussed different alignment and distribution functions used in literature. Next, we will see how attention formulations can be considerably different from each other in (i) the type of attention mechanism being used, (ii) the neural architectures, and (iii) the application domains. Figure \ref{fig:attmodel} shows the three key components of any attention modeling technique. In the remainder of this survey we will discuss a taxonomy of attention types, key neural architectures using AM and their differences, and how AM has been applied to some applications.

\section{Taxonomy of Attention}
\label{label:taxonomy}
We consider attention in four broad categories and elucidate the different types of attention within each category as shown in Table \ref{tab:term}. Note that these categories are not mutually exclusive. In fact, one can think of these categories as dimensions along which attention can be considered while employing it for an application of interest. For example, a multi-level, self and soft attention combination has been used by \cite{Yang2016HierarchicalAN}. To make this concept comprehensible, we provide a list of key technical papers and specify the multiple types of attention used within the proposed approaches in Table \ref{tab:taxonomy}.

\subsection{Number of sequences} \label{sec:num_seq}

Thus far we have only considered the case which involves a single input and corresponding output sequence. This type of attention, which we refer to as \textbf{distinctive}, is used when key and query states belong to two distinct input and output sequences respectively. Most attention models employed for translation \cite{DBLP:journals/corr/BahdanauCB14}, image captioning \cite{conf/icml/XuBKCCSZB15}, and speech recognition \cite{44926} fall within the distinctive type of attention.

A \textbf{co-attention} model operates on multiple input sequences at the same time and jointly learns their attention weights, to capture interactions between these inputs. \cite{DBLP:journals/corr/LuYBP16x} used a co-attention model for visual question answering. The authors argued that in addition to modeling visual attention on the input image, it is also important to model question attention because all words in the text of question are not equally important to the answer of the question. Further, attention based image representation is used to guide the question attention and vice versa, which essentially helps to simultaneously detect key phrases in the question and corresponding regions of images relevant to the answer. Similarly, \cite{yu2019deep} use co-attention for visual question answering task.

In contrast, for tasks such as text classification and recommendation, input is a sequence but the output is not a sequence. In this scenario, attention can be used for learning relevant tokens in the input sequence for every token in the \textit{same} input sequence. In other words, the key and query states belong to the same sequence for this type of attention. For this purpose, \textbf{self} attention, also known as inner attention has been proposed by \cite{Yang2016HierarchicalAN}. To understand this better, let's consider an input sequence of words \(\displaystyle \{w_1, w_2, w_3, w_4, w_5\} \) such that \(\displaystyle w_i \) is the vector representation of the words in the sequence. If we feed this input sequence to a self attention layer, the output is another sequence \(\displaystyle \{y_1, y_2, y_3, y_4, y_5\} \) such that \(\displaystyle y_i = \sum_j \alpha_{ij} * w_j\). Here the attention weights aim to capture how two words in the same sequence are related, where the concept of relevance depends on the main task.

\begin{table}[!h]
    \centering
    \begin{tabular}{|l|l|l|}
    \hline
         Category & Type & Characteristics \\
         \hline
         \hline
         \multirow{6}{*} { \# Sequences} & distinctive & * Key and query states are from two distinct sequences.  \\ \cline{2-3}
         & co-attention & * Multiple input sequences. \\
         &  &  * Very useful for modeling multi-modal data. \\ \cline{2-3}
         & self & * Key and query states are from the same sequence. \\
         &  & * Most popular form of attention in recent papers.\\ 
         \hline \hline
         \multirow{4}{*}{ \# Abstractions}  & single-level & * Attention is computed only once on the original input. \\ \cline{2-3}
          & multi-level & * Hierarchical attention on multiple abstractions of input. \\
          &  & * Better performance as natural hierarchy in text/images.  \\
          &  & * Computational cost increases with number of levels.  \\
         \hline \hline
         \multirow{6}{*}{ \# Positions} & soft/global & * Weighting over all positions. Computationally intensive. \\
          &  &  * Nice differentiable training objective. \\ \cline{2-3}
          &  hard & * Picks some positions by sampling from predictor.   \\
          &   & * Computationally more efficient than soft. \\
          &   & * Lacks differentiability of training objective.  \\ \cline{2-3}
          & local & * Soft-attention in a window around a position. \\
          &  & * Offers trade-off: efficient and locally differentiable.  \\
         \hline \hline
         \multirow{4}{*}{ \# Representations} & multi- & * Deals with multiple representations of the same input. \\
         & representational & * Selects representations  relevant for downstream tasks. \\ \cline{2-3}
         & multi- & * Computes relevance over each dimension of input.  \\
          & dimensional & * Extracts contextual meaning of input dimensions.\\
         \hline
    \end{tabular}
    \vspace*{0.5 cm}
    \caption{Table presents key characteristics of different types of attention within each category}
    \label{tab:term}
\end{table}
\vspace*{-0.5 cm}
\begin{table}[ht]
%\centering
\begin{tabular}{p{1.5 cm}p{2 cm}@{\hskip 0.4in}p{2 cm}p{2 cm}p{2.5 cm}p{1.5 cm}}
%\begin{tabular}{ccccc}
%\begin{tabular}{|p{0.1\linewidth}|p{0.1\linewidth}|p{0.1\linewidth}|p{0.1\linewidth}|p{0.1\linewidth}|p{0.1\linewidth}|}
\hline
\multicolumn{1}{c}{Reference} & \multicolumn{1}{c}{Application} & \multicolumn{4}{c}{Category} \\
\hline \hline
 & & Number of & Number of & Number of & Number of  \\
 & & Sequences &  Abstraction & Representations & Positions \\
 & & & Levels & & \\
\hline \hline
\cite{DBLP:journals/corr/BahdanauCB14} & Machine \newline Translation  & distinctive & single-level & - & soft \\ 
\cite{conf/icml/XuBKCCSZB15} & Image \newline Captioning & distinctive & single-level & - & hard \\ 
\cite{DBLP:journals/corr/LuongPM15} & Machine \newline Translation & distinctive & single-level & - & local \\
\cite{Yang2016HierarchicalAN} & Document \newline Classification & self & multi-level & - & soft \\
\cite{44926} & Speech \newline Recognition & distinctive & single-level & - & soft \\
\cite{DBLP:journals/corr/LuYBP16x} & Visual \newline Question \newline Answering & co-attention & multi-level & - & soft \\ 
%\cite{DBLP:journals/corr/VaswaniSPUJGKP17} & Machine Translation & & & multi-head & \\
\cite{Wang2017CoupledMA} & Sentiment \newline Classification & co-attention & multi-level & - & soft \\
\cite{Ying:2018:SRS:3304222.3304315} & Recommender \newline Systems & self & multi-level & - & soft \\
\cite{DBLP:journals/corr/abs-1709-04696} & Language \newline Understanding & self & single-level & multi-dimensional & soft \\
\cite{kiela2018dynamic} & Text \newline Representation & self & single-level &  multi-representational & soft \\ 
\hline
\end{tabular}
\vspace*{0.5 cm}
\caption{Summary of key papers for technical approaches in AMs. `-' means not applicable.}
\label{tab:taxonomy}
\vspace*{-0.5 cm}
\end{table}

\subsection{Number of abstraction levels}

In the most general case, attention weights are computed only for the original input sequence. This type of attention can be termed as \textbf{single-level}. On the other hand, attention may be applied on multiple levels of abstraction of the input sequence in a \textit{sequential} manner. The output (context vector) of the lower abstraction level becomes the query state for the higher abstraction level. Additionally, models that use \textbf{multi-level} attention can be further classified based on whether the weights are learned top-down \cite{att-via-att} (from higher level of abstraction to lower level) or bottom-up \cite{Yang2016HierarchicalAN}. 

We illustrate a key example in this category which uses the attention model at two different levels of abstraction, i.e. at word level and sentence level, for the document classification task \cite{Yang2016HierarchicalAN}. This model is called a ``Hierarchical Attention Model"(HAM) because it captures the natural hierarchical structure of documents, i.e. a document is made up of sentences and sentences are made up of words. The multi-level attention allows the HAM to extract words that are important in a sentence and sentences that are important in a document as follows. It first builds an attention based representation of sentences with first level attention applied on sequence of word embedding vectors. Then it aggregates these sentence representations using a second level attention to form a representation of the document. This final representation of the document is used as a feature vector for the classification task.

Stacked Attention Networks (SANs) proposed in \cite{10.1145/3323873.3325044} also fall into this category as they mainly employ multiple layers to iteratively refine the attention by combining information from the query (question) and results of previous attention layers. For example, the authors in \cite{10.1145/3323873.3325044} used SANs for image question answering task where multiple attention layers query the image multiple times to progressively to locate the exact regions in the image which are highly relevant for the answer. Authors claim that using global image presentation to predict the answer leads to sub-optimal results, as the attention is scattered on many objects within the first layer. But when multiple attention layers are used, higher level attention layers utilize the knowledge from lower level attention layers (visual information) and the refined query vector (question information) to extract more fine-grained and smaller regions within the image. They also observed that two attention layers are better than one, but three or more layers did not further improve the performance.

\begin{figure}[!ht]
\centering
\includegraphics[width=10cm, height=6cm]{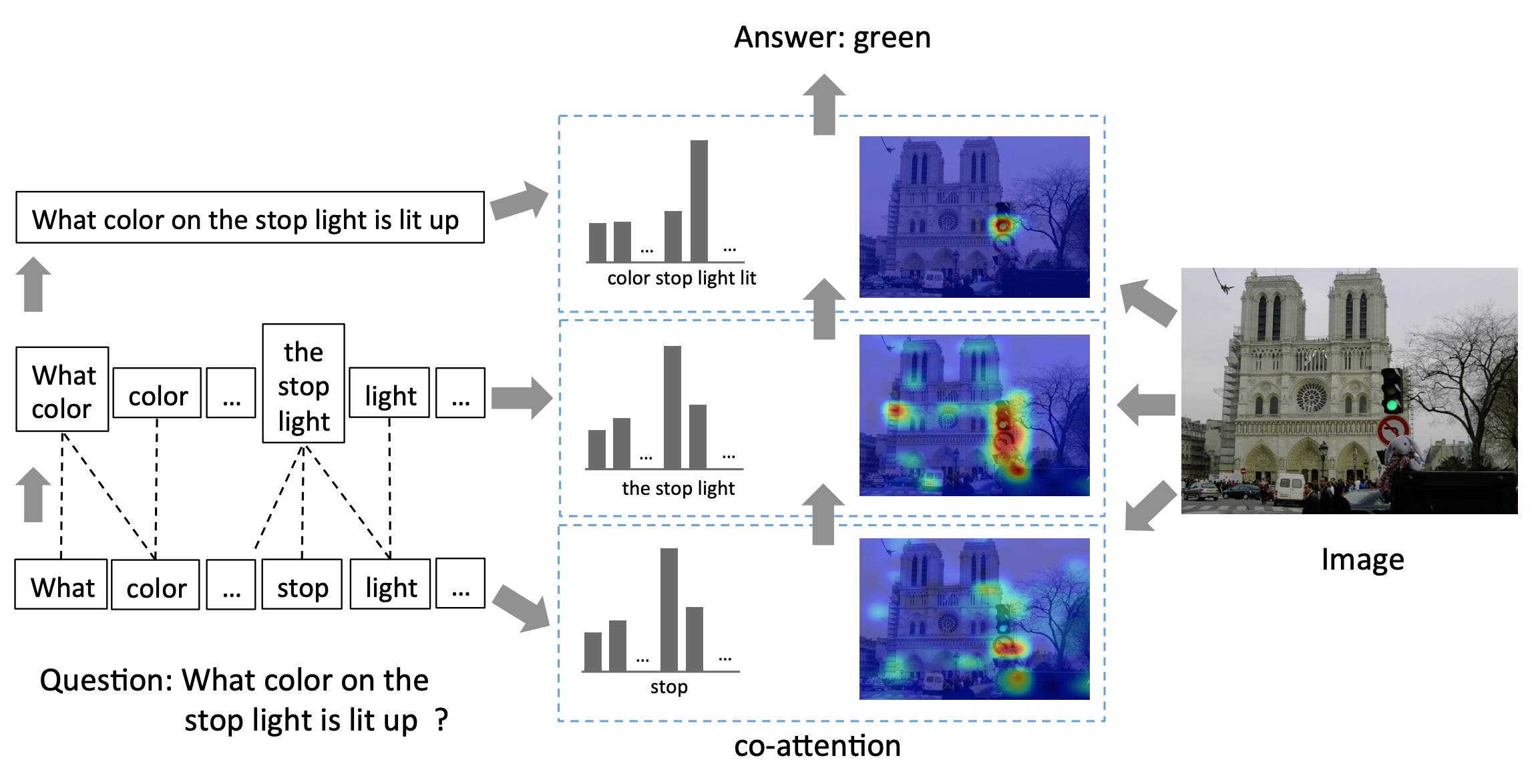} 
\caption{The AM proposed by \cite{DBLP:journals/corr/LuYBP16x} for Visual Question Answering task which is a combination of co-attention (visual and text) and multi-level (word level, phrase level and question level) attention.}
\label{co-multi}
\end{figure}

Note that the co-attention work \cite{DBLP:journals/corr/LuYBP16x} described in Section \ref{sec:num_seq} also belongs to multi-level category where it co-attends to the image and question at three levels: word level, phrase level and question level. This combination of co-attention and multi-level attention is depicted in Figure \ref{co-multi}. \cite{att-via-att} proposed ``attention-via-attention'', which uses multi-level attention (with characters on the lower level and words on the higher level) and learns the attention weights in top-down fashion.

\subsection{Number of positions}

In the third category, the differences arise from positions of the input sequence where attention function is calculated. The attention introduced by \cite{DBLP:journals/corr/BahdanauCB14} is also known as \textbf{soft} attention. As the name suggests, it uses a weighted average of all hidden states of the input sequence to build the context vector. The usage of the soft weighing method makes the neural network amenable to efficient learning through backpropagation, but also results in quadratic computational cost. %attention weight for every token in the input sequence, at a particular decoding time step is a \textit{scalar} value.

\cite{conf/icml/XuBKCCSZB15} proposed a \textbf{hard} attention model in which the context vector is computed from stochastically sampled hidden states in the input sequence. This is accomplished using a multinoulli distribution parameterized by the attention weights. The hard attention model is beneficial due to decreased computational cost, but making a hard decision at every position of the input renders the resulting framework non-differentiable and difficult to optimize. Note that these categories are not mutually exclusive. Variational learning methods and policy gradient methods in reinforcement learning have been proposed in the literature to overcome this limitation. 

\cite{DBLP:journals/corr/LuongPM15} proposed two attention models, namely \textbf{local} and \textbf{global}, in context of machine translation task. The global attention model is similar to the soft attention model. The local attention model, on the other hand, is an intermediate between soft and hard attention. The key idea is to first detect an attention point or position within the input sequence and pick a window around that position to create a local soft attention model. The position within input sequence can either be set (monotonic alignment) or learned by a predictive function (predictive alignment). Consequently, the advantage of local attention is to provide a parametric trade-off between soft and hard attention, computational efficiency and differentiability within the window. 

\subsection{Number of representations}
Generally a single feature representation of the input sequence is used in most applications. However, in some scenarios, using a single feature representation of the input may not suffice for the downstream task. In these cases, one approach is to capture different aspects of the input through multiple feature representations. Attention can be used to assign importance weights to these different representations, which can determine the most relevant aspects, disregarding noise and redundancies in the input. We refer to this model as  \textbf{multi-representational AM}, as it can determine the relevance of multiple representations of the input for downstream application. The final representation is a weighted combination of these multiple representations and their attention weights. One benefit of attention here is to directly evaluate which embeddings are preferred for which specific downstream tasks by inspecting the weights. 

\cite{kiela2018dynamic} trained attention weights over different word embeddings of the same input sentence to improve sentence representations. Similarly, \cite{maharjan2018genre} used attention to dynamically weigh different feature representations of books capturing lexical, syntactic, visual and genre information.

Based on similar intuition, in \textbf{multi-dimensional} attention, weights are induced for determining the relevance of each dimension of the input embedding vector. The intuition is that computing a score for each feature of the vector can select the features that can best describe the token’s specific meaning in any given context. This is especially useful for natural language applications where word embeddings suffer from the polysemy problem. Examples of this approach are shown in \cite{DBLP:journals/corr/LinFSYXZB17} for more effective sentence embedding representation and in \cite{DBLP:journals/corr/abs-1709-04696} for language understanding problem.

\section{Network Architectures with Attention}
\label{label:architecture}

%While Section \ref{label:taxonomy} discussed the various types of attention mechanisms that can be employed in neural architectures, in this section we cover a few critical neural network architectures which have been extensively used with attention. We classify them in two broad categories. The first category comprises of recurrent networks with sequential processing component in the network. The other category consists of network architectures which circumvent the sequential processing component of recurrent models with the use of attention mechanism. 

In this section we describe some salient neural architectures used in conjunction with attention: (1) the Encoder-Decoder framework, (2) the Transformer which circumvents the sequential processing component of recurrent models with the use of attention, (3) Memory Networks which extend attention beyond a single input sequence, and (4) Graph Attention Networks (GAT). These are some neural architectures that use AM extensively and have become popular choice in many application domains. However, exploring use of AM within various neural architectures is an active research topic, and the list of neural architectures using AM is growing fast. 

%In Table \ref{tab3} we briefly summarize all the architectures in discussion. 

%\begin{table}[H]
%\centering
%\begin{tabular}{|p{2.75cm}|p{5.25cm}|}
%\hline
%Architecture  & Description \\
%\hline
%\hline
%Encoder-Decoder & \\
%\hline
%Memory Networks & \\
%\hline
%Pointer Networks & \\
%\hline
%Transformers & \\
%\hline
%Feed Forward Networks & \\
%\hline
%\end{tabular}
%\caption{Attention Architectures}
%\label{tab3}
%\end{table}

\subsection{Encoder-Decoder}

The earliest use of attention was as part of RNN based encoder-decoder framework to encode long input sentences \cite{DBLP:journals/corr/BahdanauCB14}. Consequently, attention has been most widely used with this architecture. An interesting fact is that AM can take any input representation and reduce it to a single fixed length context vector to be used in the decoding step. Thus, it allows one to decouple the input representation from the output. One could exploit this benefit to introduce hybrid encoder-decoders, the most popular being Convolutional Neural Network (CNN) as an encoder, and RNN or Long Short Term Memory (LSTM) as the decoder. This type of architecture is particularly useful for many multi-modal tasks such as Image and Video Captioning, Visual Question Answering and Speech Recognition. 

However, not all problems where both input and output are sequential can be solved with the aforementioned formulation (e.g. sorting or travelling salesman problem). 
\textit{Pointer networks} \cite{Vinyals:2015:PN:2969442.2969540} are another class of neural models with the following two differences: (i) the output is discrete and points to positions in the input sequence (hence the name pointer network), and (ii) the number of target classes at each step of the output depends on the length of the input (and hence variable). This cannot be achieved using the traditional encoder-decoder framework where the output dictionary is known apriori (eg. in case of natural language modeling). The authors achieved this using attention weights to model the probability of choosing the i$^{th}$ input symbol as the selected symbol at each output position. This approach can be applied to discrete optimization problems such as travelling salesperson problem and sorting \cite{Vinyals:2015:PN:2969442.2969540}.

\subsection{Transformer}

Recurrent architectures rely on sequential processing of input at the encoding step that results in computational inefficiency, as the processing cannot be parallelized \cite{DBLP:journals/corr/VaswaniSPUJGKP17}. To address this, the authors in \cite{DBLP:journals/corr/VaswaniSPUJGKP17} proposed \textit{Transformer} architecture that completely eliminates sequential processing and recurrent connections. It relies \emph{only} on self attention mechanism to capture global dependencies between input and output. Authors demonstrated that Transformer architecture achieved significant parallel processing, shorter training time and higher accuracy for Machine Translation without any recurrent component. 

\begin{figure}[!h]
\centering
\includegraphics[width=8cm, height=9cm]{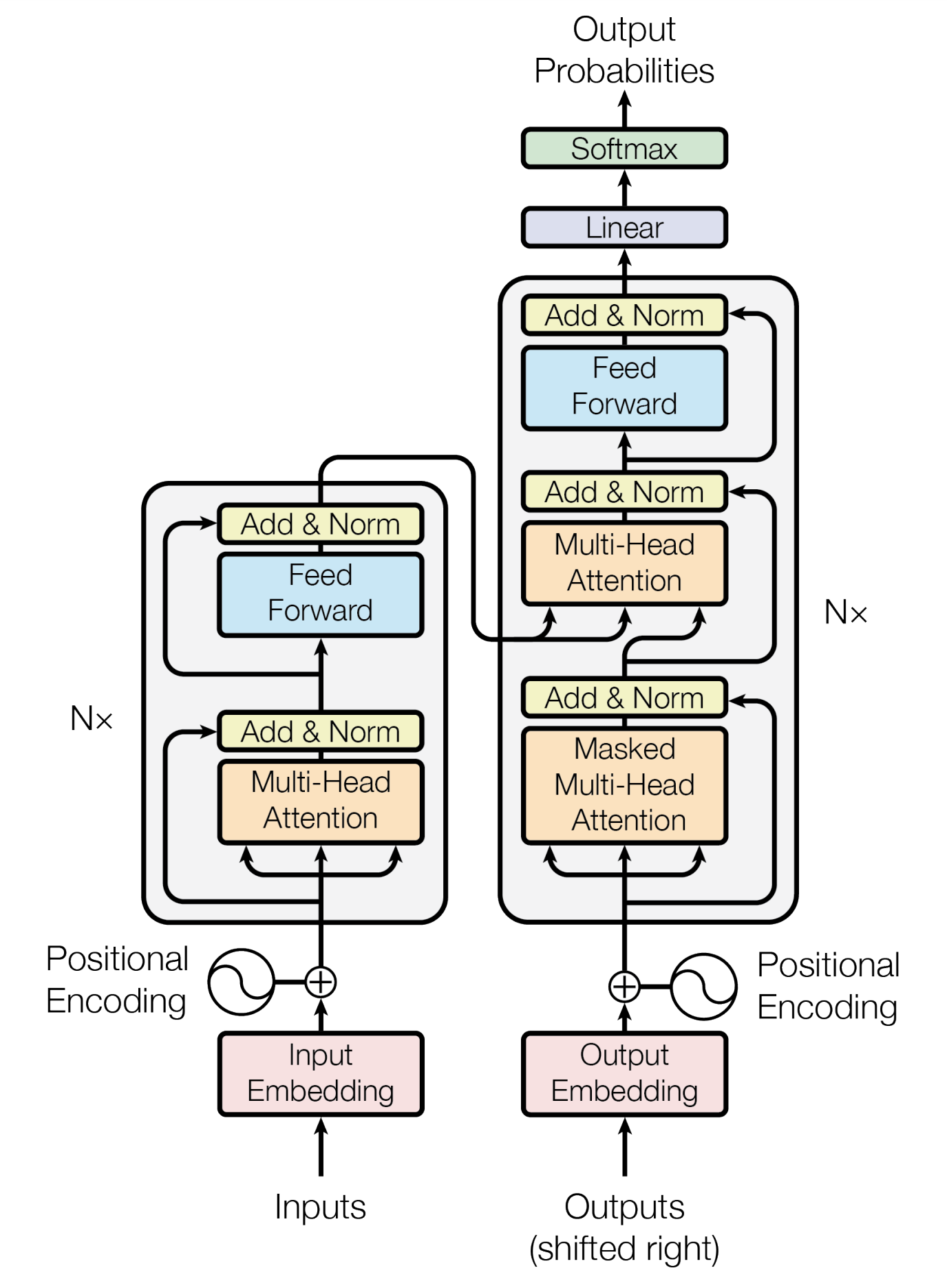} 
\caption{Transformer Architecture. Figure from \cite{DBLP:journals/corr/VaswaniSPUJGKP17}.}
\label{fig:trans}
\end{figure}

The Transformer architecture is shown in Figure \ref{fig:trans}. Transformer mainly relies on self attention mechanism which relates tokens and their positions within the same input sequence. Authors propose a novel scaled dot product alignment function for self attention mechanism, also explained in Section 3. Further, attention is known as \emph{multi-head}, because several attention layers are stacked in parallel, with different linear transformations of the same input. In other words, rather than only computing the attention once, the multi-head mechanism splits the input into fixed-size segments and then computes the scaled dot-product attention over each segment in parallel. The independent attention outputs are then concatenated into expected dimensions. The main architecture is composed of a stack of 6 identical layers of encoders and decoders with two sub-layers: point-wise Feed Forward Network(FFN) layer and multi-head self attention layer. Point-wise FFN means same linear transformation is applied to each position in the sequence independently, increasing parallel processing. The decoder is similar to the encoder, except that the decoder contains two multi-head attention sub-modules instead of one. The first multi-head attention sub-module is masked to prevent positions from attending to the future. One additional feature of this architecture is usage of positional encoding. The positional encoding is used because input is sequential which demands the model to make use of the temporal aspect of the input (order information), yet components that capture this positional information (i.e. RNNs / CNNs) are not used. To account for this, the encoder phase in the Transformer generates content embedding as well as positional encoding for each token of the input sequence. Finally, normalization and residual connections are mechanisms used to help the model train faster and more accurately.

Transformers can capture global/long range dependencies between input and output, support parallel processing, require minimal inductive biases (prior knowledge), demonstrate scalability to large sequences and datasets, and allow domain-agnostic processing of multiple modalities (text, images, speech) using similar processing blocks. Consequently, Transformer architecture has become state-of-the-art approach for many mainstream NLP, Computer Vision and Cross-Modal tasks. Moreover, there has been an increasing interest in the use of attention models in a wide variety of applications after Transformer, making this architecture a significant milestone for attention. The number of variants of Transformer and its applications have grown to the extent that individual surveys have been published on this topic eg. \cite{khan2021transformers}. We describe some key variants of Transformer and its applications for NLP, Computer Vision and Cross-Modal tasks in Section 6.1, 6.2 and 6.3 respectively.

However, although Transformer is undoubtedly a huge improvement over the RNN based sequential models, it comes with its own share of limitations: i) input text has to split into fixed number of segments resulting in context fragmentation ii) high
parametric complexity which results in computational cost and resources iii) large training data requirements due to minimal inductive bias iv) difficulty in interpreting what self attention mechanism learns and what is the contribution of input tokens towards predictions. Consequently, multiple lines of work have been established to introduce improvements in Transformers. 

One direction of research analyzes the multi-head self attention in Transformers. \cite{clark-etal-2019-bert} perform analysis of Transformer based (pre-trained) language model called Bidirectional Encoder Representations(BERT) to determine that particular heads correspond well to particular relations such as direct objects of verbs, determiners of nouns, objects of prepositions etc. Further, first layer consists of high-entropy heads that produce bag-of-vector-like representations by having broader attention span. \cite{voita-etal-2019-analyzing} evaluate the contribution made by attention heads in the encoder to the overall performance of the model and find that: i) vast majority of heads can be pruned without affecting performance ii) important heads have specialized and linguistically interpretable functions in the model. \cite{kobayashi-etal-2020-attention} perform norm-based analysis of transformed input vectors to reveal that contrary to previous studies, BERT pays poor attention to special tokens, and word alignment can be extracted from attention mechanisms of Transformer. Lastly, \cite{NEURIPS2019_2c601ad9} find that many attention layers can even be individually reduced to a single attention head without affecting performance, increasing its efficiency and propose a novel technique for pruning.

Another line of work aims to improving attention span is to make the context that can be used in self-attention longer, more efficient and flexible. Transformer-XL by \cite{dai-etal-2019-transformer} is one such work that solves the context fragmentation problem by adopting new positional encoding and reusing hidden states between segments. Another work by \cite{sukhbaatar-etal-2019-adaptive} propose a novel self attention mechanism that can learn its optimal attention span. This allows to extend the context size used in Transformer, while maintaining the computational cost. Finally, since the computational cost of Transformer grows quadratically with sequence length. Hence, an important research direction is to reduce the computation time and memory consumption. Sparse Transformers \cite{DBLP:journals/corr/abs-1904-10509}, Reformers \cite{DBLP:conf/iclr/KitaevKL20} and Performers \cite{choromanski2020rethinking} are some recent approaches towards this end.
%Attention can only deal with fixed-length text strings. The text has to be split into a certain number of segments or chunks before being fed into the system as input. This chunking of text causes context fragmentation. For example, if a sentence is split from the middle, then a significant amount of context is lost. In other words, the text is split without respecting the sentence or any other semantic boundary

% This helps the model to capture various aspects of the input and improves its expressiveness.
  %Within the last 3 years, multiple variants of Transformer have been adopted for a wide variety of problems such as OpenAI's GPT and GPT-2 (decoder-only transformers for Language Modeling) \cite{noauthororeditor}, Bidirectional Encoder Representations from Transformer(BERT) for Language Representations \cite{devlin-etal-2019-bert}, Image Transformer for Image Generation \cite{pmlr-v80-parmar18a}, Universal Transformer for Question Answering \cite{DBLP:conf/iclr/DehghaniGVUK19} and Reinforcement Learning \cite{DBLP:journals/corr/abs-1910-06764} to name a few.

\subsection{Memory Networks}

%Inspiration to add external memory to neural networks can be attributed to the power of Turing machines and its use of an infinite memory. Technically, ``memory'' is sort of built into the weights of a network as it learns different filters (in CNNs) or in the states (in RNNs and LSTMs). However these typically are not able to remember or memorize inputs from the past. 

Applications like Question Answering (QA) and chat bots require the ability to learn from information in a database of facts. The input to the network is a knowledge database and a query, where some facts are more relevant to the query than others. In this case, an external memory is needed to store the knowledge database of facts and attention is crucial to selectively focus on the relevant facts. It can be considered as a generalization of attention, wherein instead of modeling attention over a single sequence, it is used over a large database of sequences (facts). We consider three approaches in the literature which couple an external memory component with attention for Question Answering namely, End-to-End Memory Networks \cite{sukhbaatar2015end}, Dynamic Memory Networks \cite{pmlr-v48-kumar16}, and Neural Turing Machines \cite{DBLP:journals/corr/GravesWD14}. We can think of memory networks as generally having three components: (i) A process that “reads” raw database, and converts them into distributed representations. (ii) A list of feature vectors storing the output of the reader. This can be understood as a “memory” containing a sequence of facts, which can be retrieved later, not necessarily in the same order, without having to visit all of them. (iii) A process that “exploits” the content of the memory to sequentially perform a task, at each time step having the ability put attention on the content of one memory element (or a few, with a different weight).

\begin{figure}[!h]
\centering
\begin{subfigure}[b]{0.45\textwidth}
\includegraphics[width=6.5cm, height=5.5cm]{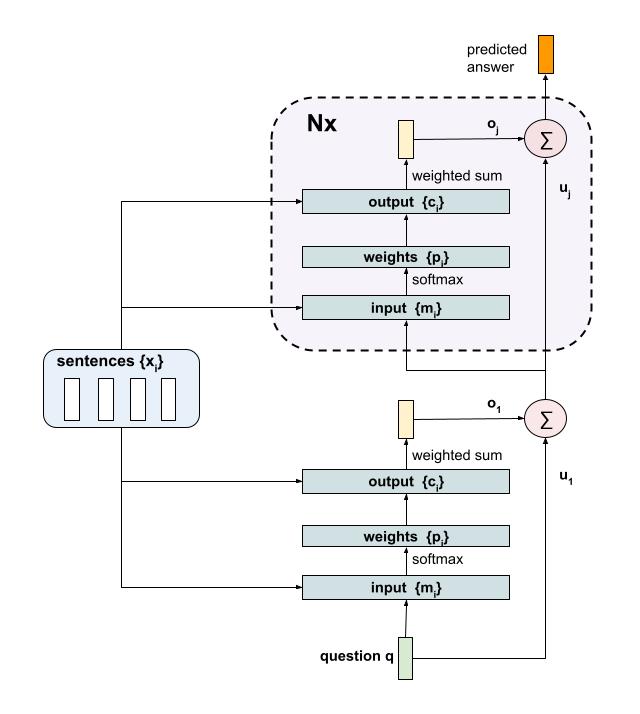} 
\caption{End-to-End Memory Network Architecture}
\end{subfigure}
\hspace*{0.5cm}
\begin{subfigure}[b]{0.45\textwidth}
\includegraphics[width=6.5 cm, height=5cm]{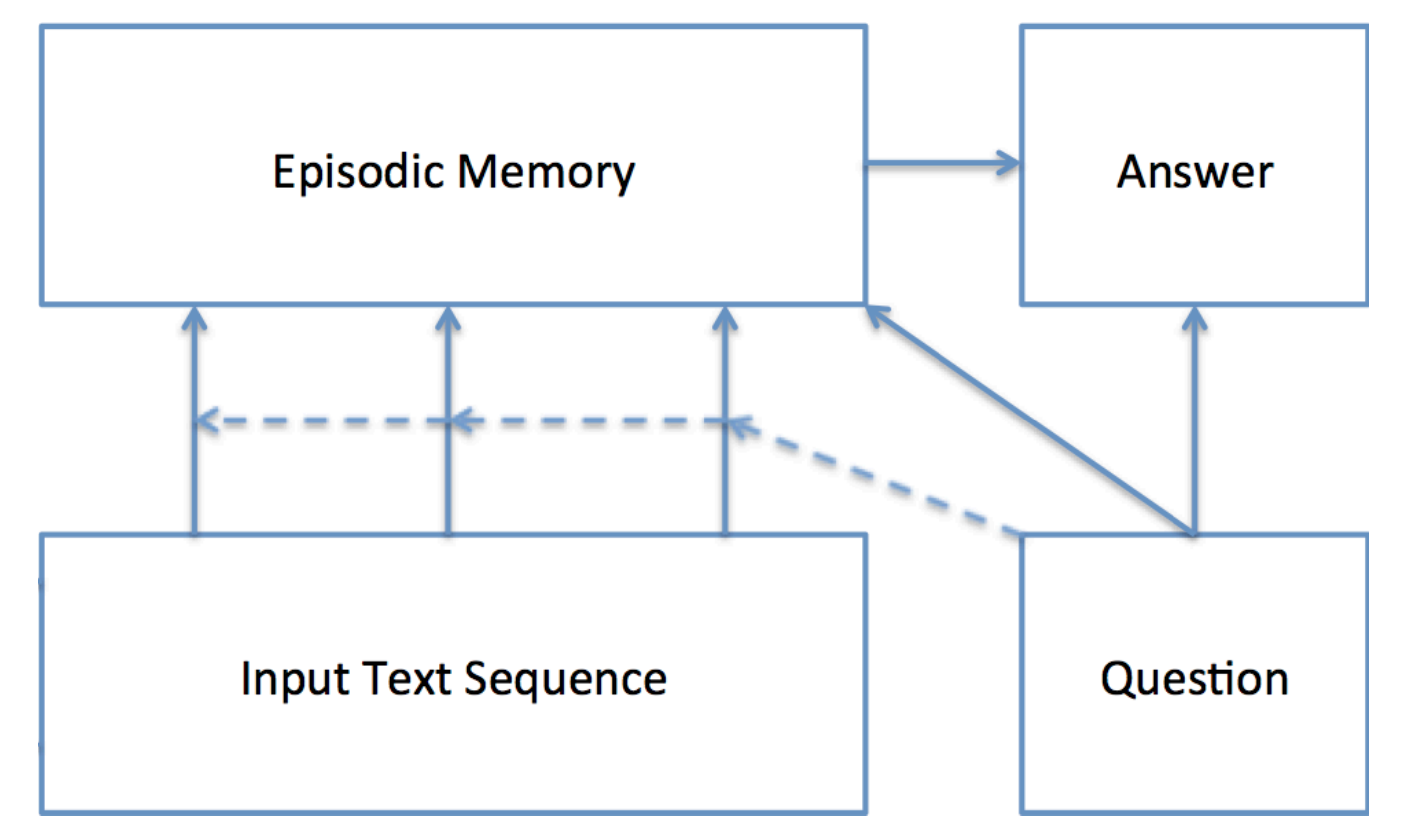}
\caption{Dynamic Memory Network Architecture Overview. Figure from \cite{pmlr-v48-kumar16}}
\end{subfigure}
\caption{Memory Network Architectures}
\label{fig:imagemm}
%\vspace*{-0.5 cm}
\end{figure}

End-to-End Memory Networks (MemN2N) derive its name from the fact that they enable end-to-end training via backpropagation compared to Memory Networks (MemNN) \cite{DBLP:journals/corr/WestonCB14} and require less supervision, making them applicable to wider range of NLP tasks such as Question Answering and Language Modeling. It also allows multiple reads ("hops") over the long-term memory before generating output token which is crucial for good performance these tasks. The architecture shown in Figure \ref{fig:imagemm}(a) can be broken down into two main parts. First part of the architecture helps to find the relevant facts for the query from the knowledge database using inner product between query and each memory vector, followed by softmax operation, to find the best match. In the second part, final answer for the query is calculated using the context vector over relevant facts with the help of attention. 

Dynamic Memory Networks (DMN) use an episodic memory module, which chooses which parts of the inputs to focus on through the attention mechanism and outputs a memory vector representation. It repeats this process iteratively by conditioning the attention over the question as well as the previous memory representation, which allows the module to: i) attend to different inputs during each iteration ii) retrieve additional information, which was thought to be irrelevant in previous iterations. Figure \ref{fig:imagemm}(b) shows an overview of the architecture. Questions trigger gates which control whether certain input are given to the episodic memory module. The final state of the episodic memory (after multiple episodes/iterations) is the input to the answer module. Another work by \cite{10.5555/3045390.3045643} demonstrates the use of Dynamic Memory Networks to answer questions based on images. The input module extracts feature vectors from images using a CNN based network which are then fed to the episodic memory module.

%Questions trigger an iterative attention process which allows the model to condition its attention on the inputs and the result of previous iterations. These results are then reasoned over in a hierarchical recurrent sequence model to generate answers.

%Overview of DMN modules. Communication between them is indicated by arrows and uses vector representations.

%The iterative nature of this module allows it to attend to different inputs during each pass. It also allows for a type of transitive inference, since the first pass may uncover the need to retrieve additional facts. 

%achieve this by using an array of memory blocks to store the database of facts, and using attention to model relevance of each fact in the memory for answering the given query. 
%Using attention also provides computational advantage by making the objective continuous and End-to-End Memory Networks can 

%Although more computational hops improve performance, this model was unable to beat Memory networks trained with strong supervision. Furthermore, the soft memory lookup is hard to scale as the size of the memory increases. 

%This multi-hop attention was shown to be essential to good performance, and the model was able to retrieve and reason about several supporting facts to answer a specific question.

Neural Turing Machine (NTM) also uses a continuous (albeit smaller) memory representation along with a controller (typically feed-forward network or LSTM) that dictates read/write operations on the memory. The system is similar to a Turing Machine but is differentiable end-to-end, allowing it to be efficiently trained with gradient descent. It uses attention to access the memory selectively, constraining the read and write operation to interact with a small portion of the memory and making interactions with the memory highly sparse. However, the NTM memory uses both content and address-based access and is used to infer simple algorithms such as copying, sorting and associative recall from input and output examples.

The biggest advantage of memory networks is that it can store information as memory and use it effectively and scalably via attention mechanism by focusing on only small, relevant part of the memory. MemN2N have shown superior performance on tasks such as Question Answering and Language Modeling compared to RNNs and LSTMs. DMNs have shown state-of-the-art results on Sentiment Analysis and Part of Speech Tagging.

\subsection{Graph Attention Networks (GAT)}

Application domains such as social networks, citation networks, protein-protein interactions, cheminformatics, etc. come naturally in the form of graph-structured data. Arbitrary graphs have a varying number of neighbors and it is therefore considerably harder to work with them compared to sequences and images. 
In fact, one could think of sequences and images as special form of graphs with highly rigid and regular connectivity pattern (each element in sequence is connected to two adjacent elements; each pixel in an image is connected to its eight neighbouring pixels).

\begin{figure}[!h]
\centering
\includegraphics[width=12cm, height=7cm]{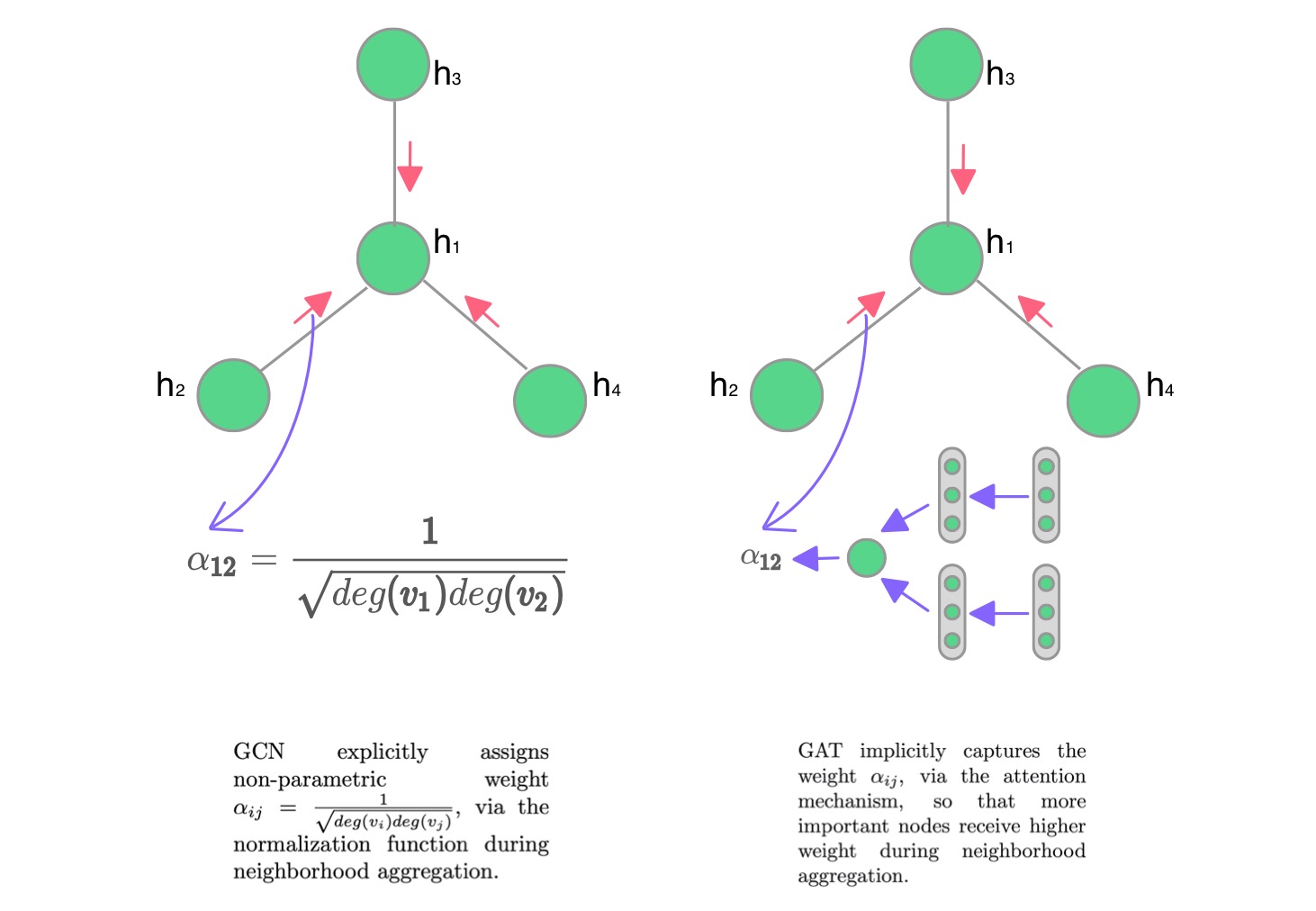} 
\caption{Neighbor importance in Graph Convolutional Network vs Graph Attention Network. \newline Figure from https://dsgiitr.com/blogs/gat/.}
\label{fig:gat}
\end{figure}

Generalizing convolutional layers used for images to graph convolutional layers would thus require innovations to enable:
(i) computational and storage efficiency - model should not require more than O(V+E) time and memory, (ii) fixed number of parameters - the number of parameters should not depend on the input graph size else the model becomes unmanageable, (iii) localisation - the model should be able to work independently on a local neighbourhood of a node to enable parallel computation and scalability, (iv) ability to specify arbitrary importance to different neighbours so as to correctly incorporate relevance of a neighbor for each node and not treat all neighbors equally or based on some structural property such as node degree, and (v) applicability to arbitrary, unseen graph structures.

Graph Convolutional Network (GCN) showed good performance on node classification task by combining local graph structure and node-level features. However, as shown in the Figure \ref{fig:gat}, GCN assigns explicit non-parametric weight to neighbors based on their node degree.
\cite{Velickovic2018GraphAN} propose Graph Attention Networks (GATs) that employ self-attention over the node features of neighbors so that more important nodes receive higher weight. GATs are computationally efficient as the operation of the self-attention layer can be parallelized. No eigendecomposition or computationally intensive matrix operations are required. Moreover, GATs allow for assigning different importance to
nodes of a same neighborhood via attention weights, enabling a higher model capacity than GCNs.
Finally, the attention mechanism is applied in a shared manner to all edges in the graph, and therefore it does not depend on upfront access to the global graph structure.

The real-world graph usually comes with multi-types of nodes and edges, also widely known as heterogeneous graphs. Heterogeneous graphs contain more comprehensive information and richer semantics, therefore are more useful for data mining tasks. However, due to this complexity of heterogeneous graphs, homogeneous graph approaches cannot be directly applied to heterogeneous graphs.
\cite{wang2019heterogeneous} extend graph attention for heterogeneous
graphs using hierarchical attention. In particular, given the node features as input, a transformation matrix projects different types of node features into the same space. Then the node-level attention is able to learn the attention values between the nodes and their meta-path (connections defined on multiple nodes) based neighbors, while the semantic-level attention aims to learn the attention values of different meta-paths for the specific task in the heterogeneous graph. Based on the learned attention values in terms of the two levels, the model can get the optimal combination of neighbors and multiple meta-paths in a hierarchical manner, which enables the learned node embeddings to better capture the complex structure and rich semantic information. The overall model is optimized via backpropagation in an end-to-end manner.

\begin{table}
    \centering
    \begin{tabular}{p{2.5 cm}|p{3.5 cm}|p{6.5 cm}}
    \hline
         Application Domain & Application & Seminal Works \\
         \hline
         \hline
         \multirow{6}{*}{\shortstack[l]{Natural Language \\ Processing}} & Machine Translation & \cite{8948335}, \cite{DBLP:journals/corr/abs-1808-08946}, \cite{DBLP:journals/corr/VaswaniSPUJGKP17}, \cite{DBLP:journals/corr/BritzGLL17}, \cite{DBLP:journals/corr/BahdanauCB14}, \cite{DBLP:journals/corr/LuongPM15} \\ \cline{2-3}
         & Summarization & \cite{xu-etal-2020-self}, \cite{nallapati-etal-2016-abstractive}, \cite{chopra-etal-2016-abstractive}, \cite{DBLP:journals/corr/RushCW15} \\ \cline{2-3}
         & Text Classification and Representation & \cite{kiela2018dynamic}, \cite{DBLP:journals/corr/LinFSYXZB17}, \cite{Yang2016HierarchicalAN} \\ \cline{2-3}
         %& Pre-trained Language Models &  \\ \cline{2-3}
         & Sentiment Analysis & \cite{wang-etal-2020-relational}, \cite{D16-1058}, \cite{Ma2018TargetedAS}, \cite{DBLP:journals/corr/TangQL16}, \cite{DBLP:journals/corr/abs-1812-07860} \\ \cline{2-3}
         & Question Answering & \cite{DBLP:conf/iclr/DehghaniGVUK19}, \cite{DBLP:journals/corr/HermannKGEKSB15}, \newline \cite{sukhbaatar2015end} \\ \cline{2-3}
         & Pre-trained Language Models & \cite{lewis-etal-2020-bart}, \cite{Lan2020ALBERT}, \cite{NEURIPS2020_1457c0d6}, \cite{NEURIPS2019_dc6a7e65}, \cite{dai-etal-2019-transformer}, \cite{devlin-etal-2019-bert} \\
         \hline \hline
         \multirow{4}{*}{Computer Vision} & Image Classification & \cite{dosovitskiy2021an}, \cite{touvron2021training}, \cite{Zhao_2020_CVPR},  \cite{Jetley2018LearnTP}, \cite{10.5555/2969033.2969073} \\ \cline{2-3}
         & Image Generation & \cite{pmlr-v119-chen20s}, \cite{pmlr-v80-parmar18a}, \cite{DBLP:journals/corr/GregorDGW15},  \\ \cline{2-3}
         & Object Detection & \cite{DBLP:journals/corr/abs-2010-04159}, \cite{DBLP:conf/eccv/CarionMSUKZ20}, \cite{DBLP:conf/iclr/LinWPSSDJA20}, \cite{ba2014multiple} \\ \cline{2-3}
         & Image Synthesis & \cite{zhang2019self} \\
         \hline \hline
         \multirow{4}{*}{\shortstack[l]{Multi-Modal \\ Tasks}} & Multimedia (Image, Video) Description & \cite{wang-etal-2020-cross-media}, \cite{Sun_2019_ICCV}, \cite{conf/icml/XuBKCCSZB15}, \cite{10.1109/ICCV.2015.512}, \cite{DBLP:journals/corr/ChoCB15} \\ \cline{2-3}
         & Speech Recognition & \cite{8977559}, \cite{44926}, \cite{Chorowski:2015:AMS:2969239.2969304} \\ \cline{2-3}
         & Visual Question \newline Answering & \cite{NEURIPS2019_c74d97b0}, \cite{DBLP:conf/emnlp/TanB19}, \cite{li2019visualbert}, \cite{anderson2018bottomup}, \cite{DBLP:journals/corr/LuYBP16x},  \\ \cline{2-3}
         & Human Communication Comprehension & \cite{zadeh2018multi} \\ 
         \hline \hline
         \multirow{2}{*}{\shortstack[l]{Recommender \\ System}} & User Profiling &  \cite{articleZhuAAAI19}, \cite{NAIRS}, \cite{He2018NAISNA}, \cite{DBLP:journals/corr/abs-1711-06632}, \cite{DBLP:journals/corr/abs-1808-09781}, \cite{ijcai2018-546} \\ \cline{2-3}
          & Item/User Representations & \cite{ijcai2020-415}, \cite{deng-etal-2020-hierarchical}, \cite{wu-etal-2019-hierarchical-user} \\ \cline{2-3}
          & Exploit Auxiliary Information & \cite{DBLP:journals/datamine/XiaoZPSZY21}, \cite{DBLP:journals/tkde/WuCHFXW20}, \cite{10.1145/3292500.3330989} \\ \cline{2-3}
         \hline \hline
         \multirow{4}{*}{\shortstack[l]{Graph-based \\ Systems}} & Graph Classification & \cite{10.1145/3219819.3219980}, \cite{10.1145/3097983.3098088} \\ \cline{2-3}
         & Graph to Sequence \newline Generation & \cite{DBLP:conf/aaai/ZhengFW020}, \cite{xu2018graph2seq} \\ \cline{2-3}
         & Node Classification & \cite{10.1145/3394486.3403092}, \cite{Velickovic2018GraphAN}, \newline  \cite{NIPS2018_8131}, \cite{feng-etal-2016-gake} \\ \cline{2-3}
         & Hyperedge Detection & \cite{Zhang2020Hyper-SAGNN} \\ 
         %& Link Prediction & \cite{10.5555/3172077.3172382} \\
         \hline
    \end{tabular}
    \vspace*{0.25 cm}
    \caption{Summary of key applications of AMs}
    \label{tab:app}
\end{table}

\section{Applications}
\label{label:applications}
%: why attention, nature of AM, unique challenges of domain, representative techniques
Attention models have become an active area of research because of their intuition, versatility and interpretability. Variants of attention models have been used to address unique characteristics of a diverse set of application domains. In some applications, attention models have shown a significant impact on the performance for the task at hand, whereas in others they have helped to learn better representations of entities such as documents, images and graphs. In some cases, attention has entirely transformed the field of application by becoming the most popular choice of technique. A few such examples are Machine Translation, pre-trained embeddings with BERT and Question Answering.

Given that the areas of application are very broad, in this work, we mainly discuss the need for attention models for each application domain, a few instances of applications within each domain and cover their seminal work in Table \ref{tab:app} which can become a starting point for further investigation. We describe attention modeling in the following application domains: (i) NLP, (ii) Computer Vision, (iii) Multi-Modal Tasks, (iv) Recommender Systems and (v) Graphical Systems.

\subsection{Natural Language Processing(NLP)}
In the NLP domain, attention assists in focusing on the relevant parts of the input sequence, alignment of input and output sequences, and capturing long range dependencies for longer sequences. For instance, modeling attention in neural techniques for \textbf{Machine Translation} allows for better alignment of sentences in different languages, which is a crucial problem in MT. This automatic alignment of sentences in different languages helps to capture subject-verb-noun locations which differ from language to language. The advantage of the attention model also becomes more apparent while translating longer sentences \cite{DBLP:journals/corr/BahdanauCB14}. The longer the sentence, the harder it is to embed all the content and alignment information in the vanilla technique without attention. Several studies including \cite{DBLP:journals/corr/BritzGLL17} and \cite{DBLP:journals/corr/abs-1808-08946} have shown performance improvements in Machine Translation using attention. \cite{8948335} have presented GRU-gated attention model for Machine Translation. The GRU-gating mechanism is useful to avoid computation of similar context vectors at different decoding steps, allowing them to be more discriminatory in nature.

\textbf{Question Answering} problems have made use of attention to better understand questions by focusing on relevant parts of the question \cite{DBLP:journals/corr/HermannKGEKSB15} and store large amount of information using memory networks to help find answers \cite{sukhbaatar2015end}. Another seminal work by \cite{DBLP:journals/corr/RushCW15} made significant advancement in abstractive sentence \textbf{summarization} task by using soft attention mechanism. Such data driven approaches had proven to be challenging in the past for the task of summarization but the proposed method showed significant gains compared to several existing baselines. \cite{xu-etal-2020-self} uses a Transformer based model to enhance the copy mechanism in abstractive summarization. The self attention in Transformer is used to guide the copy mechanism so that it can focus on important words in the source text which need to be extracted into the summary.

In the \textbf{Sentiment Analysis} task, self attention helps to focus on the words that are important for determining the sentiment of input. A couple of approaches for aspect based sentiment classification by \cite{wang-etal-2020-relational}, \cite{D16-1058} and \cite{Ma2018TargetedAS} incorporate additional knowledge of aspect related concepts into the model and use attention to appropriately weigh the concepts apart from the content itself. Sentiment Analysis application has also seen multiple architectures being used with attention such as Memory Networks \cite{DBLP:journals/corr/TangQL16} and Transformer \cite{DBLP:journals/corr/abs-1812-07860}.

Other applications within the NLP domain which have employed attention models extensively include Text Classification, and Text Representation Learning. As mentioned earlier in Section \ref{label:taxonomy}, \textbf{Text Classification} and \textbf{Text Representation} problems mainly make use of self attention to build more effective sentence or document representations/embeddings. \cite{Yang2016HierarchicalAN} use a multi-level self attention, whereas \cite{DBLP:journals/corr/LinFSYXZB17} propose a multi-dimensional, and \cite{kiela2018dynamic} propose a multi-representational self attention model. 

Lastly, many applications of NLP have been completely revolutionised with the advent of \textbf{Pre-Trained Language Models}. Pre-trained language models have proven extremely beneficial due to the following reasons: i) they are trained on large corpus that can learn universal language representations capturing many facets of language such as long-term dependencies, hierarchical relations, and sentiment ii) they can be easily fine tuned for multiple downstream NLP tasks with significantly less labeled data, avoiding training a new model from scratch iii) they democratize development of NLP applications by allowing easier model building. Transformer based pre-trained language models have been especially popular recently with three main types: i) Transformer and its variants such as Transformer-XL \cite{dai-etal-2019-transformer}, BART \cite{lewis-etal-2020-bart} ii) Bidirectional Encoder Representations from Transformers (BERT) \cite{devlin-etal-2019-bert} and its variants such as RoBERTa \cite{DBLP:journals/corr/abs-1907-11692}, ALBERT \cite{Lan2020ALBERT} iii) Generative Pre-trained Transformer (GPT) and its variants such as GPT-2 \cite{noauthororeditor}, GPT-3 \cite{NEURIPS2020_1457c0d6}. XLNet combines BERT and Transformer-XL \cite{NEURIPS2019_dc6a7e65}.

Although Transformer has been used for several NLP tasks, Transformer-XL understands context beyond the fixed-length limitation of Transformer and can learn 450\% longer dependency, critical to achieve better performance on both short and long sequences. BART is similar to Transformer in architecture but is trained to reconstruct the original text from corrupted input text with an arbitrary noising function. On the other hand, BERT uses bi-directional encoder such that it allows the model to consider the context from both the left and the right sides of each word. RoBERTa improves over BERT by using a larger dataset for training, training the model over more iterations, and removing the next sequence prediction training objective. ALBERT uses parameter reduction techniques such as factorized embedding parameterization, cross-layer parameter sharing to reduce the number of parameters by 18x and faster training by 1.7x. Finally XLNet combines the capabilities of BERT and Transformer-XL to achieve state-of-the-art performance on 18 NLP tasks including question answering, natural language inference, sentiment analysis, and document ranking.

OpenAI's GPT is an unsupervised language model trained on a giant collection of free text corpora. GPT specifically uses multi-layer decoder only Transformer as well as does not generate embeddings for usage in downstream NLP tasks but fine-tunes the base model itself. The successor to GPT and GPT-2, GPT-3 is similar to GPT-2, but it uses alternating dense and sparse attention patterns as in Sparse Transformer \cite{DBLP:journals/corr/abs-1904-10509}. This large scale transformer-based language model has been trained on 175 billion parameters and has shown strong performance for over two dozen NLP tasks. GPT-3 is the largest model so far, and its impressive capabilities have positioned it to outrank other pre-trained models.

\subsection{Computer Vision(CV)}

Visual attention has become popular in many main stream CV tasks to focus on relevant regions within the image, and capture structural long-range dependencies between parts of the image. Visual attention term was conceived by \cite{10.5555/2969033.2969073} in which attention was proposed for the \textbf{Image Classification} task. Here the authors use attention to not only select relevant regions and locations within the input image but also to reduce computational complexity of CNNs by processing only selected regions at high resolution. This is crucial to control the computational complexity of proposed model irrespective of the size of the input image, compared to CNNs whose computational complexity scales linearly with increase in the size of the image (number of image pixels). 

Visual attention also provides a significant benefit for \textbf{Object Detection}, as it can aid to localize and recognize objects within the image. In \cite{ba2014multiple} authors use attention for multiple object detection problem where the image is processed in a sequential manner ("glimpse" at a time) to learn to predict one object at a time. Hence, a sequence of labels is generated in the end for multiple objects, until there are no more objects that the model can recognize. Deep Recurrent Attentive Writer (DRAW) \cite{DBLP:journals/corr/GregorDGW15} exploit attention for \textbf{Image Generation}. Although it is an encoder-decoder framework which compresses and regenerates images during training; the major difference from previous work is it generates images in step by step fashion, rather than in a single pass. This is accomplished by using attention to selectively attend to parts of the input image while regenerating specific scenes within the image in an iterative manner. Self-Attention Generative Adversarial Networks (SAGANs) \cite{zhang2019self} use a self-attention mechanism into convolutional GANs by calculating the response at a position as a weighted sum of the features at all positions. This helps in capturing long range dependencies efficiently compared to convolution processes alone, as they process the information in a local neighborhood. The self-attention module is complementary to convolutions by modeling long range, multi-level dependencies across image regions efficiently.

Although CNNs have become the dominant models for vision tasks since 2012, CNNs are designed specifically for images and can be computationally demanding. For next generation of efficient, scalable and domain agnostic architectures, using \textbf{Transformers} for vision tasks has become a new research direction. First major work in this direction has been the Vision Transformer (ViT) by \cite{dosovitskiy2021an}, which directly uses the original Transformer architecture on image patches along with positional embeddings for image classification task. It has outperformed a comparable state-of-the-art CNN with four times fewer computational resources. Another work by \cite{touvron2021training} uses novel distillation approach for Transformers to perform large-scale image classification, without pre-training on an external large dataset, making it more efficient than ViT. Similarly, Detection Transformer (DETR) by \cite{DBLP:conf/eccv/CarionMSUKZ20} is the first Object Detection framework which relied on Transformers as the main building block. DETR uses the flattened output from CNNs and positional encodings as input to directly generate output classes and bounding boxes. It also streamlined the pipeline by removing many hand designed components (eg. anchor generation, non-maximum suppression) required in existing approaches. \cite{DBLP:journals/corr/abs-2010-04159} further use deformable attention module to reduce the training time of DETR by 10X, lowering its computational cost. Finally, Transformers have also been used for Image Generation task with Image Transformer by \cite{pmlr-v80-parmar18a} and Image GPT by \cite{pmlr-v119-chen20s}, designed to sequentially predict each pixel of an output image given its previously generated pixels.

%Overall, there are two major model architectures for adopting Transformers in vision: i) pure transformer architecture, ii) hybrid architecture which combines CNN and Transformer. 

%The state-of-the-art approaches in convolutional Generative Adversarial Networks (GANs) excel at synthesizing image classes which are distinguished more by texture than by geometry. For example, they are good for landscape classes such as sky and ocean, but fail to capture geometric or structural patterns that occur in other classes. As an example, dogs are often drawn with realistic fur texture but without clearly defined separate feet) \cite{zhang2019self}. Convolution processes the information in a local neighborhood, thus using convolutional layers alone is computationally inefficient for modeling long-range dependencies. To address the issues due to long range dependencies, 
\subsection{Multi-Modal Tasks}
Attention has been extensively used for multi-modal applications because it helps to understand relationships between different modalities. \textbf{Multimedia Description} is the task of generating a natural language text description of a multimedia input sequence which can be image and video \cite{DBLP:journals/corr/ChoCB15}. Similar to QA, here attention performs the function of finding relevant parts of the input image \cite{conf/icml/XuBKCCSZB15} to predict the next word in caption or focus on smaller subset of frames for video description \cite{10.1109/ICCV.2015.512}. %Further, \cite{Li2017MAMRNNMA} exploit the temporal and spatial structures of videos using multi-level attention for video captioning task. The lower abstraction level extracts specific regions within a frame and higher abstraction level focuses on small subset of frames selectively.

For \textbf{Speech Recognition}, in \cite{44926} the authors claim that without attention, the model significantly overfits the data because it memorizes the training transcripts, without really paying attention to the acoustics. \cite{Chorowski:2015:AMS:2969239.2969304} also describe how Speech Recognition differs from other NLP and CV tasks as the input is much noisier, lacks a clear structure and has multiple similar speech fragments. In this work, authors propose an attention mechanism such that it takes into account both the location and content of the important fragments in the input sequence. Adapting the attention mechanism to also incorporate location helps with longer input sequences and recognition of similar/repeated speech fragments. Latest work by \cite{8977559} propose a novel approach for visually inspecting the encoder representations so that they can be used to understand how attention mechanism works for speech recognition task.

Another interesting work on \textbf{Human Communication Comprehension} \cite{zadeh2018multi} addresses the challenging problem of comprehending face-to-face communication which is a complex multi-modal task involving language, vision and speech modalities simultaneously. Here attention is specifically used for discovering interactions between different modalities (called cross-view dynamics) at each time step. The authors demonstrated that the approach shows state-of-the-art performance on multiple tasks such as speaker trait recognition and emotion recognition. 

Again, Transformers have also been used extensively for vision-language tasks to learn generic representations that can effectively encode cross-modal relationships. The two main types of Transformers used in this domain are: single and multi stream \cite{khan2021transformers}. In single stream models such as videoBERT \cite{Sun_2019_ICCV} and visualBERT \cite{li2019visualbert}, all multi-model inputs are given to a single Transformer as input, that automatically discovers relationships between the two domains. On the other hand, ViLBERT \cite{NEURIPS2019_c74d97b0} and LXMERT \cite{DBLP:conf/emnlp/TanB19} fall in the multi-stream category, where independent Transformers for used for each modality first and later cross-modal representations are learned using another co-attentional Transformer.

%and (iii) improve performance in visual QA task by modeling multi-modality in input using co-attention \cite{DBLP:journals/corr/LuYBP16x}. 

\subsection{Recommender Systems}
Attention has seen a significant usage in recommender systems for user profiling, learning user/item representations, and exploiting auxiliary information such as knowledge graph, social network. User profiling aims to assign attention weights to interacted items of a user to capture long and short term interests in a more effective manner. This is intuitive because all interactions of a user are not relevant for the recommendation of an item and user’s interests are transient as well as varied in the long and short time span. Attention mechanism has been used for finding the most relevant items in user's history to improve recommendations in Collaborative Filtering \cite{NAIRS, He2018NAISNA} as well as RNN based sequential models \cite{DBLP:journals/corr/abs-1808-09781, ijcai2018-546, articleZhuAAAI19, DBLP:journals/corr/abs-1711-06632}. 

Learning effective user and item representations lies at the heart of recommender systems. Consequently, \cite{ijcai2020-415} use attention for cross domain recommendation problem, in which attention is used to effectively combine user/item embeddings learned from both domains, to generate a single embedding for every user/item. Similarly, \cite{deng-etal-2020-hierarchical} use HAN for learning more effective item representations for paper review rating recommendation problem. Specifically, authors use hierarchical attention to leverage the hierarchical structure of the paper reviews with three levels: sentence (level one), intra-review (level two) and inter-review (level three). Similarly, \cite{wu-etal-2019-hierarchical-user} also use three-tier HAN to to learn user and item representations from reviews as different reviews, sentences and even words have different informativeness for modeling users and items.

Attention is also helpful to utilize auxiliary information in recommender systems more effectively. For example, \cite{10.1145/3292500.3330989} propose Knowledge Graph Attention Network (KGAT) in which a hybrid graph linking user-item interactions and item attributes is constructed to exploit higher order relations. A node's embedding is computed by learning attention weights over its neighbours which can consist of users, items or attributes. \cite{DBLP:journals/tkde/WuCHFXW20} use HAN to attend to three key aspects that affect user's preference for image recommendation: upload history, social influence and owner admiration. Attention is used at two levels: element level while learning individual aspect representations and aspect level since not all aspects are equally important for learning user's preferences. Finally, recent work by \cite{DBLP:journals/datamine/XiaoZPSZY21} proposes a social recommendation model which explores higher-order friends of user in a social network to help content recommendation. The proposed method allow a user’s context vector to attend to friend's context vectors so that interests and knowledge aggregated from a user's social network can be utilized for recommendation.

\subsection{Graph-based Systems}

%Graph-structured data arise naturally in many different application domains. By representing data as graphs, we can capture entities (i.e., nodes) as well as their relationships (i.e., edges) with each other. Many interesting and important problems have been studied in this area. These include graph classification \cite{duvenaud2015convolutional}, link prediction \cite{sun2011co}, community detection \cite{girvan2002community}, node classification and clustering \cite{perozzi2014deepwalk}.

Many important real-world datasets come in the form of graphs or networks; examples include social networks, knowledge graphs, protein-interaction networks, and the World Wide Web. Attention has been used to highlight elements of the graph (nodes, edges, subgraphs) which are more relevant for the main task \cite{lee2018attention}. The common approach is to compute attention-guided embeddings of nodes or edges or subgraphs or their combinations. Attention architecture in graphs is efficient, since it is parallelizable across node neighbor pair, can be applied to graph nodes having different degrees, and is directly applicable to inductive learning problems, including tasks where the model has to generalize to completely unseen graphs. In contrast to Graph Convolutional Networks, attention mechanism in graphs allows for assigning different importance to nodes of a same neighborhood, enabling a leap in model capacity. Analyzing the learned attention weights may lead to benefits in interpretability. Attention has been used in several machine learning tasks in graph-structured data including (i)  \textbf{Node Classification} \cite{Velickovic2018GraphAN, 10.1145/3394486.3403092}, (ii) \textbf{Link Prediction} \cite{10.5555/3172077.3172382}, (iii) \textbf{Graph Classification} \cite{10.1145/3219819.3219980}, and (iv) \textbf{Graph to Sequence} Generation \cite{DBLP:conf/aaai/ZhengFW020, xu2018graph2seq}.
\cite{Zhang2020Hyper-SAGNN} uses attention for hyperedge prediction in hypergraphs. Hypergraphs are mainly used to represent higher order interactions in graph using hyperedges i.e. edges connecting multiple nodes. Most existing methods are designed for analyzing pair-wise interactions and thus are unable to effectively capture higher-order interactions in graphs. Consequently, authors propose self attention based GAT for hyperedge prediction which can work with both homogeneous and heterogeneous hypergraphs with variable hyperedge size.

\begin{figure}[!h]
\centering
%\begin{subfigure}[b]{0.3\textwidth}
%\includegraphics[width=4cm, height=4cm]{alignment.png} 
%\caption{Alignment of French and English sentences in MT %\cite{DBLP:journals/corr/BahdanauCB14}}
%\end{subfigure}
%\hspace*{0.25cm}
\begin{subfigure}[b]{0.45\textwidth}
\includegraphics[width=5.5 cm, height=5cm]{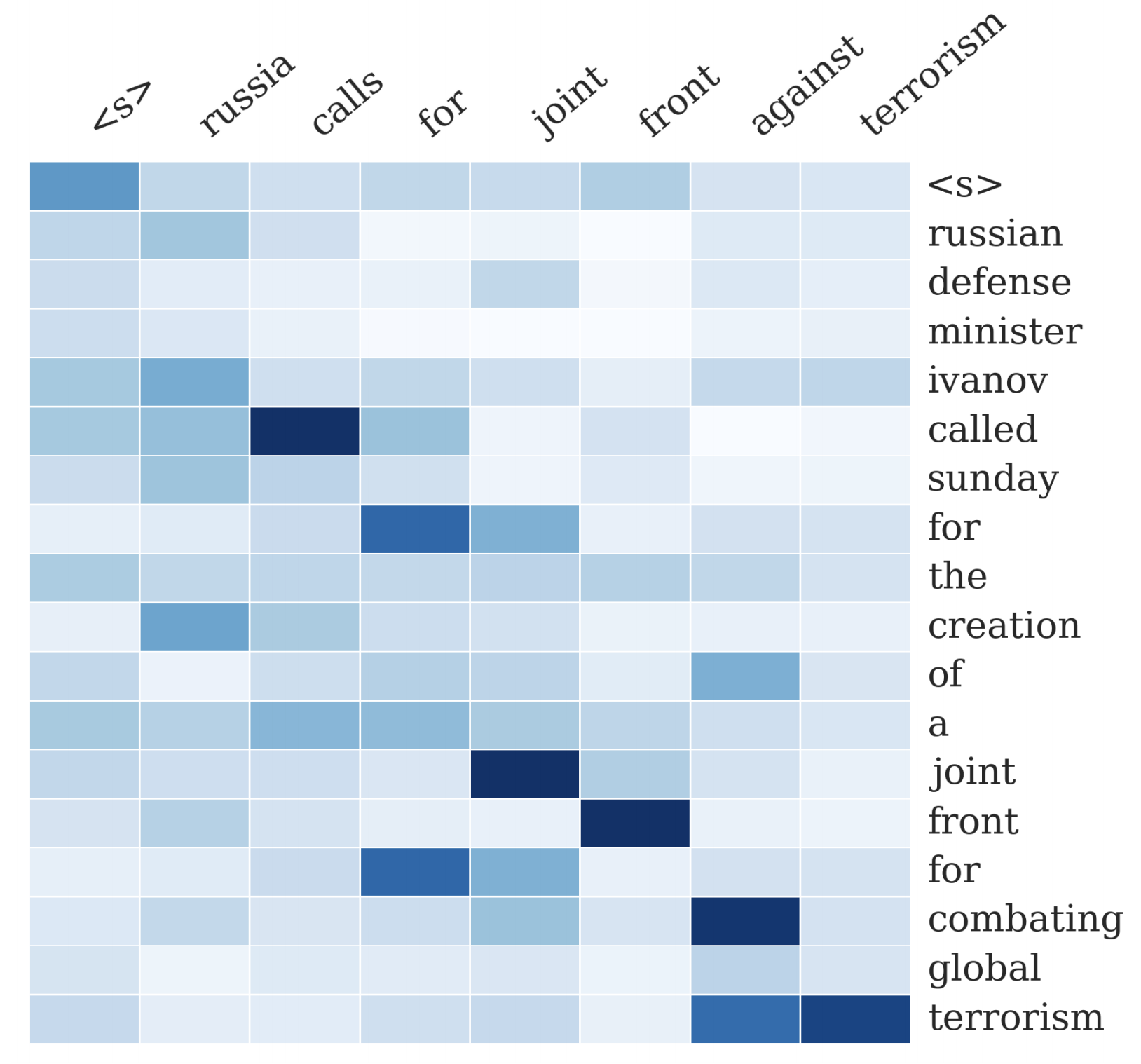}
\caption{Alignment of input and output sequences for summarization. Figure from \cite{DBLP:journals/corr/RushCW15}.}
\end{subfigure}
\hspace*{0.4cm}
\begin{subfigure}[b]{0.45\textwidth}
\includegraphics[width=5.5 cm, height=5cm]{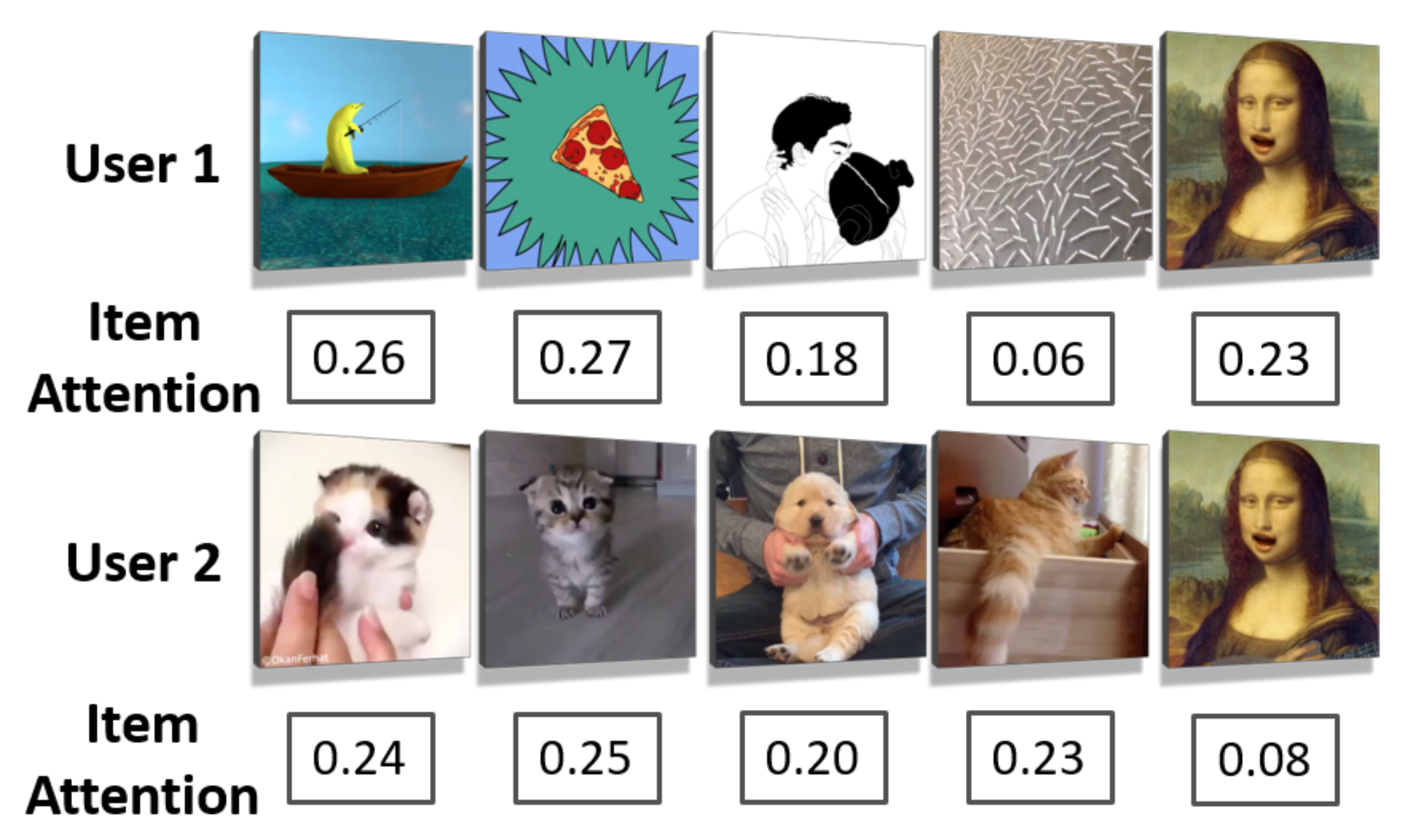}
\caption{Weights of items in user's history for recommendation. Figure from \cite{He2018NAISNA}.}
\end{subfigure}
\begin{subfigure}[b]{0.95\textwidth}
\centering
\vspace{0.5cm}
\includegraphics[width=12cm, height=6cm]{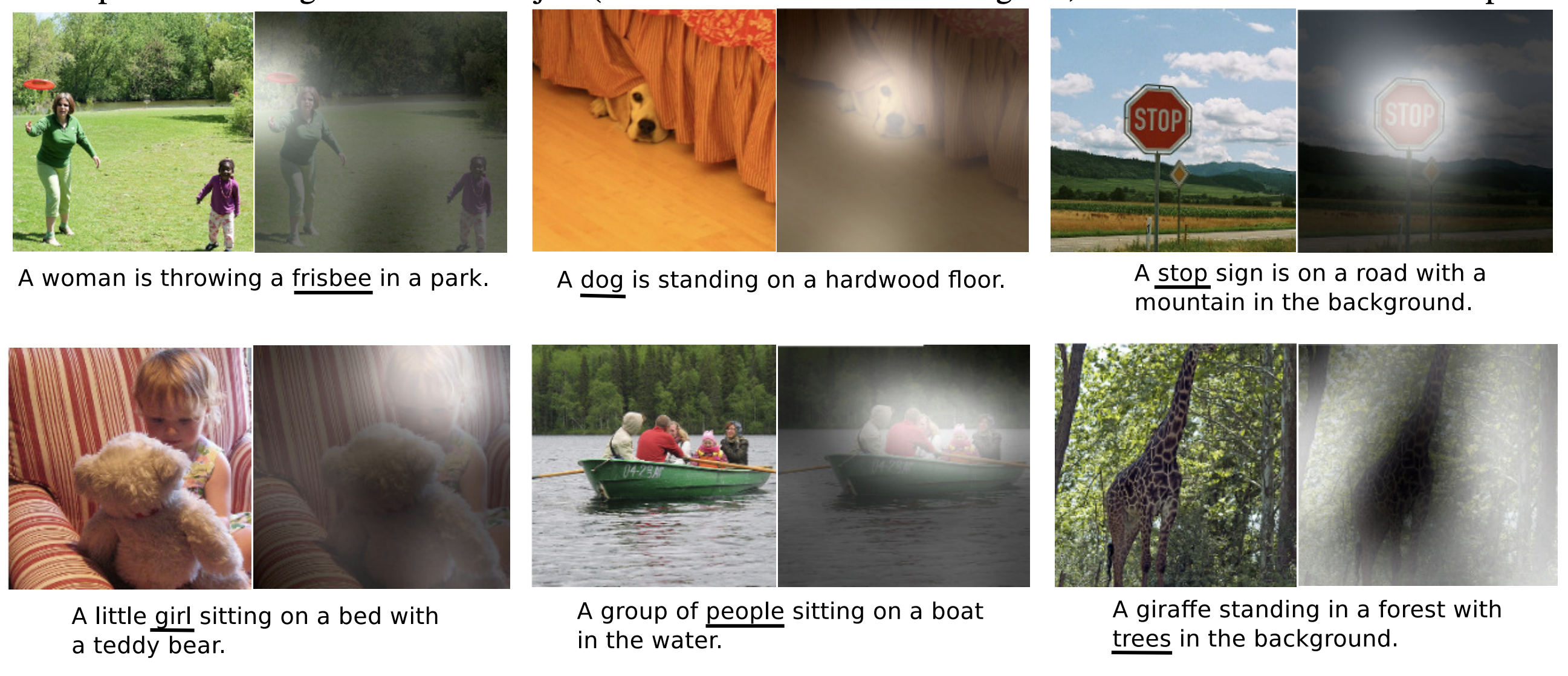} 
\caption{Relevant image regions for image captioning. Figure from \cite{conf/icml/XuBKCCSZB15}.}
\end{subfigure}
\caption{Examples of visualization of attention weights.}
\label{fig:image3}
\vspace*{-0.5 cm}
\end{figure}

\section{Attention for Interpretability}
\label{label:interpretability}

There is a growing interest in the interpretability of AI models - driven by both performance as well as transparency and fairness of models\footnote{https://fatconference.org}. However, neural networks, particularly deep learning architectures have been criticized for their lack of interpretability \cite{DBLP:journals/corr/abs-1802-01933}. 

Modeling attention is particularly interesting from the perspective of interpretability because it allows us to directly inspect the internal working of the deep learning architectures. The hypothesis is that the magnitude of attention weights correlates with how relevant a specific region of input is for the prediction of output at each position in a sequence. This can be easily accomplished by visualizing the attention weights for a set of input and output pairs. \cite{DBLP:journals/corr/LiMJ16a} upholds attention as one of the important ways to explain the inner workings of neural models.

As shown in Figure \ref{fig:image3}(a), \cite{DBLP:journals/corr/RushCW15} showed that attention model is able to focus on relevant words in the input sequence while generating output for the summarization task. In the given example, the input word \textit{combating} has a high attention weight for the output word \textit{against} which demonstrates that attention model can capture word relationships for summarization. Figure \ref{fig:image3}(b) shows attention weights can help to recognize user's interests. User 1 seems to have a preference for “cartoon” videos, while user 2 prefers videos on “animals” \cite{He2018NAISNA}. \cite{conf/icml/XuBKCCSZB15} provide extensive list of visualizations of the relevant image regions (i.e. with high attention weights) which had a significant impact on the generated text in the image captioning task (example shown in Figure \ref{fig:image3}(c)). Similarly, \cite{DBLP:journals/corr/BahdanauCB14} visualize attention weights which clearly show automatic alignment of sentences in French and English despite the fact that subject-verb-noun locations differ from language to language. In particular, attention model shows non-monotonic alignment by correctly aligning \textit{environnement marin} with \textit{marine environment}. 
%As shown in Figure \ref{fig:image3}(a), \cite{DBLP:journals/corr/BahdanauCB14} visualized attention weights for some input sequences to show clear alignment between certain parts of the input and output text in MT task. They observed that although alignment was largely monotonic with larger weights along the diagonal, attention model also learned a number of non-trivial, non-monotonic alignments eg. nouns and adjectives are ordered differently in French and English but model was still able to correctly identify these correlated positions. 

We also summarize a few other interesting findings as follows. \cite{de2019bias} explored gender bias in occupation classification, and showed how the words getting more attention during classification task are often gendered. \cite{Yang2016HierarchicalAN} noted that the importance of words \textit{good} and \textit{bad} is context dependent for determining the sentiment of the review. The authors inspected the attention weight distribution of these words to find that they span from 0 to 1 which means the model captures diverse context and assign context-dependent weight to the words. \cite{44926} noted that in speech recognition, attention between character output and audio signal can correctly identify start position of the first character in audio signal and attention weights are similar for words with acoustic similarities. Finally, \cite{kiela2018dynamic} found that the multi-representational attention assigned higher weights to GloVe, FastText word embeddings out of many other representations used, particularly GloVe for low frequency words. As another interesting application of attention, \cite{D17-2021} and \cite{D18-2007} provide a tool for visualizing attention weights of deep-neural networks. The goal is to interpret and perturb the attention weights so that one can simulate what-if scenarios and observe the changes in predictions interactively.

Despite being popularly used to shed light on inner working of black-box neural networks, using attention weights for model explainability remains an area of active research. Some articles have presented a contradictory viewpoint that challenges the usage of attention weights as explanations of model behaviour/decision making process \cite{jain-wallace-2019-attention}, \cite{serrano-smith-2019-attention}. Based on several experiments on application of attention models for NLP tasks, \cite{jain-wallace-2019-attention}  argued that attention weights are often not correlated with the typical feature importance analysis. Moreover, they performed two analyses to observe the sensitivity of predictions to the change in attention weights and observed that changing attention weights with random permutations and adversarial training do not change the output predictions. \cite{serrano-smith-2019-attention} applied a different analysis based on intermediate representation erasure method and showed that attention weights are at best noisy predictors of relative importance of the specific regions of input sequence, and should not be treated as justifications for model's decisions.

\section{Conclusion}
\label{label:conclusion}

In this survey we discussed different ways in which attention has been formulated in the literature, and attempted to provide an overview
of various techniques by discussing a taxonomy of attention, key neural network architectures using attention, and application domains that have seen significant impact. We discussed how the incorporation of attention in neural networks has led to significant gains in performance, provided greater insight into neural network's inner working by facilitating interpretability, and also improved computational efficiency by eliminating sequential processing of input. We hope that this survey provides an understanding of the different directions in which research has been done on this topic, and how techniques developed in one area can be applied to other domains. We conclude this survey with some of the emerging research directions in attention modeling.

\subsection{Real-time Attention}

In a usual neural machine translation model, encoding and decoding of the entire sentence happens sequentially. However, some real-time applications such as live video captions or conversations between people speaking different languages demand that machine translation model starts generating a translation before it has finished reading the entire source sentence. \cite{chiu2017monotonic} use monotonic chunkwise attention that adaptively split the input sequence into smaller chunks over which soft attention is computed, thus allowing online and linear-time decoding. \cite{ma2019monotonic} enable online decoding with Transformers by using monotonic multi-head attention which alternates between encoding and decoding. With an ever-increasing demand for real-time applications, we expect online attention to be an important area for future research.

\subsection{Stand-alone Attention}

Introducing attention in state-of-the-art models such as CNNs in Computer Vision has shown performance gains. The question is whether attention can be a stand-alone primitive for vision models instead of serving as an augmentation on top of convolutions. \cite{ramachandran2019stand} investigated stand-alone self-attention in vision models by replacing all instances of spatial convolutions with self-attention over local regions and found that these pure self-attention vision models are able to compete with state-of-art models on benchmark vision datasets. \cite{wang2020axial} enables performing attention within a larger or even global region by factorizing 2D self-attention into two 1D self-attentions.

\subsection{Model Distillation}

A number of industry applications such as recommender and search systems have strict latency constraints for online model serving. While pre-trained models such as BERT have shown considerable performance improvements, they have hundreds of millions of parameters which makes them inapplicable for online serving. The field of model distillation aims to compress an existing large, complex model with a simpler model while retaining its accuracy. \cite{wang2020minilm} used deep self-attention to train a small model (student) by deeply mimicking the self-attention module of the large model (teacher). Moreover, they observed that introducing a teacher assistant (a concept introduced in \cite{mirzadeh2020improved}) also helps the distillation of large pre-trained Transformer models. Similarly, \cite{touvron2020training} employed a teacher-student strategy specific to
transformers ensuring that the student
learns from the teacher through attention.

\subsection{Attention for Interpretability}

Exploring the relationship between attention weights and model interpretability continues to be an active area of research. Future research can investigate attention distributions of current models and how they can be modified to offer plausible justifications of model prediction. An example of such work is by \cite{mohankumar2020towards} where they observe that in LSTM based encoders the hidden representations at different time-steps are very similar to each other; even a random permutation of the attention weights does not affect the model’s predictions. They propose a diversity-driven LSTM cell which uses an orthogonalization technique to ensure that the hidden states are farther away from each other in their spatial dimensions. These modified LSTM cells generate attention weights that (i) provide a more precise importance ranking of the hidden states (ii) are better indicative of words important for the model’s predictions (iii) correlate better with gradient-based attribution methods. 

\subsection{Auto-learning Attention}

The automated design of neural network architectures using neural architecture search (NAS) has outperformed human designs on various tasks. An open question is if we can use NAS to search the optimal architecture of high
order attention module.
\cite{ma2020auto} is the first attempt to extend NAS to search plug-and-play attention modules beyond the
backbone architecture. They define a novel concept of attention module named Higher Order Group Attention that
can represent high order attentions and utilize a differentiable search method to search the optimal attention module efficiently.

\subsection{Multi-instance Attention}

Existing attention mechanisms attend to individual items in the memory with a fixed granularity, e.g., a word token or a pixel in an image grid. Multi-instance attention is a generalization that allows attending to structurally adjacent group of items, e.g., 2D areas in images, or subsequences in natural language sentences. One of the first techniques working with attention over multiple instances is \emph{area attention} (\cite{li2019area}). A simple approach is to model the key of an area as the mean vector of the key of each item in the area. Alternatively, a richer representation of each area can be formed by using derived features such as standard deviation of the key vectors within each area.
Such formulations can be very useful in exploring attention mechanism on groups of items of dynamic shapes and sizes.

\subsection{Multi-agent Systems}

Understanding and modeling behavior of multi-agent systems is required in many real-world applications, including autonomous vehicles, multi-player games, etc. Attention mechanism working with deep generative models can be used to model interactions within multi-agent systems.  
\cite{li2020generative, fujii2020policy} use attention to capture behavior generating process of multi-agent systems, as well as identify agent groups and how they interact with each other.

\subsection{Scalability}

Large Transformer models have shown extraordinary success in achieving state-of-the-art results in many NLP and computer vision applications. However, training and deploying these models can be prohibitively costly for long sequences (such as in bioinformatics) as the standard self-attention mechanism of the Transformer uses \(\displaystyle O(n^2) \) time and space with respect to sequence length. An important theme of research is to reduce the quadratic time and space complexity of Transformers to linear without loss in performance of these models. \cite{sukhbaatar2019adaptive} propose an approach that uses a learnable masking function to dynamically control attention span in self attention mechanism. This allows to increase the maximum context size used in Transformers, without increasing computational cost. \cite{choromanski2020rethinking} introduce Performers, which approximate softmax full-rank attention in linear space and time complexity. \cite{wang2020self} demonstrate that the self-attention mechanism can be approximated by a low-rank matrix and exploit this finding to propose a new self-attention mechanism, which reduces the overall self-attention complexity from quadratic to linear in both time and space. The resulting linear transformer, the Linformer, performs on par with standard Transformer models.  \cite{li2020sac} propose an \emph{Edge Predictor} which utilizes an LSTM model to dynamically predict attention edges (relationship between tokens), thus automatically searching for the best attention patterns in long sequences.

\bibliographystyle{ACM-Reference-Format}
\bibliography{tist-acmsmall}

\end{document}